\documentclass{article}

\usepackage[final]{corl_2026}
\usepackage{amsmath}
\usepackage{amsfonts}
\usepackage{graphicx}
\usepackage{float}
\usepackage{booktabs}
\usepackage{amssymb}
\usepackage{wrapfig}
\usepackage{media9}
\usepackage{animate}
\usepackage{multirow}
\usepackage{cleveref}
\usepackage[most]{tcolorbox}
\usepackage{placeins}
\usepackage{hyperref}
\usepackage{longtable}
\usepackage{array}
\definecolor{mydarkblue}{rgb}{0,0.08,0.45}
\graphicspath{{images/}}

\title{RMMBench: A Comprehensive Benchmark\\for Robotic Mobile Manipulation}

\author{
  Huapeng Li\textsuperscript{1}, Fuxiang Feng\textsuperscript{1}, Jinqiu Fan\textsuperscript{1}, Shuo Yang\textsuperscript{1}, \vspace{2pt}\\ 
  \vspace{2pt} \textbf{Fengjiao Chen\textsuperscript{2}, Xuezhi Cao\textsuperscript{2}, 
  Ran Song\textsuperscript{1*}, Wei Zhang\textsuperscript{1}}\\
  \textsuperscript{1}School of Control Science and Engineering, Shandong University \quad \textsuperscript{2}Meituan Group
}

\begin{document}
\maketitle
\renewcommand{\thefootnote}{\fnsymbol{footnote}}
\footnotetext[1]{Corresponding author: Ran Song $<$\texttt{ran.song@sdu.edu.cn}$>$.}
\setcounter{footnote}{0}

\begin{abstract}
Although the advancement of vision-language models (VLMs) has endowed robots with enhanced environmental understanding and task reasoning, a comprehensive evaluation methodology is important to advance the integration of VLMs in robotic navigation and manipulation. However, current benchmarks lack a comprehensive method to evaluate diverse robotic tasks, and evaluation metrics remain relatively constrained, making it difficult to assess the embodied capabilities of VLMs in a thorough and fine-grained manner. To address this issue, we propose RMMBench, an evaluation benchmark that requires robots to understand language instructions and perform long-horizon tasks in continuous spaces. RMMBench seamlessly integrates high- and low-level embodied tasks into a unified framework, constructing a ``navigation-manipulation'' task suite comprising 90 canonical task scenarios that range from localized manipulation to long-horizon composite navigation. The results reveal that leading VLMs still face major challenges in spatial localization when performing mobile manipulation tasks, and also highlight the necessity of enhancing the spatial perception capability of robots during long-horizon interactions. RMMBench can be accessed at \url{https://mxxq-stack.github.io/rmmbench-project/}.
\end{abstract}

\keywords{Mobile Manipulation, Benchmark, VLM, Robot Learning}

\begin{figure}[H]
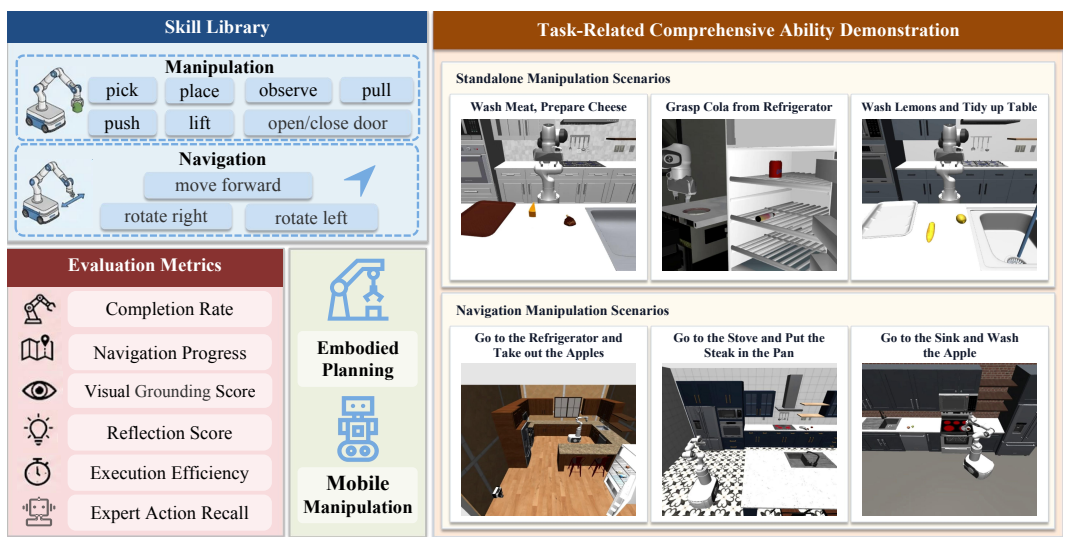

    \centering
\animategraphics[autoplay,loop,width=1\linewidth]{10}{images/abstract_v1/abstract_v1_}{1}{20}
\vspace{-7pt}
    \caption{\textbf{Overview of the RMMBench.} RMMBench is a comprehensive mobile manipulation benchmark with a standardized skill library to evaluate the embodied performance of vision-language models. Please use Adobe Acrobat Reader to view embedded video demonstrations.
    }
    \vspace{7pt}
    \label{fig:rmmbench_overview}
\end{figure}

\section{Introduction}
The integration of natural language instructions to guide robotic navigation and manipulation tasks represents an ongoing pursuit in the robotics community \cite{openai2024gpt}. Traditional learning-based policies and task and motion planning frameworks initially establish a classical pipeline that decouples high-level reasoning from low-level execution \cite{ho2016generative, andrew2018reinforcement, garrett2021integrated, ren2024extended, durante2024agent}. Recently, large language models (LLMs) and vision-language models (VLMs) have been widely adopted as high-level planners for robot learning \cite{brown2020language, huang2023voxposer, zitkovich2023rt,Zhao-RSS-24}. This shift has led to the development of various benchmarks designed to evaluate these models in robotic tasks.

Early benchmarks focused primarily on discrete symbolic tracking and high-level abstract task planning \cite{jiang2023vima}, limiting the output of a model to macro-commands or semantic goals. To test the proficiency of a model in handling diverse tasks, recent work introduced multi-stage, long-horizon benchmarks featuring various scenarios \cite{nasiriany2024robocasa, zhang2025vlabench}. However, to isolate semantic reasoning or simplify evaluation pipelines, such benchmarks commonly employed encapsulated interaction middleware or separate vision components (such as SAM \cite{ravi2025sam} or Grounding DINO \cite{liu2024grounding}) to provide spatial grounding \cite{gao2025genmanip, li2025emmoe, yang2025embodiedbench}.

Due to the inherent difficulty of alignment in navigation-to-manipulation transitions, existing benchmarks for more complex mobile manipulation tasks typically relied on underlying navigation modules for path finding and resorted to heuristic-based grasping \cite{li2025emmoe, yang2025embodiedbench}. However, this simplification fails to rigorously test the autonomous navigation capability of the models and their precision in execution. Consequently, they often relied on coarse-grained, outcome-oriented metrics (e.g., task success rates) for performance evaluation, leaving the continuous tracking of intermediate action execution unexamined. Most benchmarks characterized tasks through macro-level concepts such as ``semantic understanding'' or ``commonsense reasoning'' \cite{zhang2025vlabench, gao2025genmanip, yang2025embodiedbench}. Without fine-grained execution metrics, the actual quality of the generated low-level actions remains unquantified, failing to provide detailed diagnostic insights. Therefore, a comprehensive benchmark based on raw visual inputs and continuous execution is required to analyze the capability of a model in low-level spatial reasoning and mobile manipulation.

To address such evaluation requirements and establish a more comprehensive diagnostic paradigm, this work introduces RMMBench, an evaluation benchmark for grounded spatial perception and composite mobile manipulation. Table~\ref{tab:benchmark_comparison} compares RMMBench with some popular benchmarks for embodied intelligence. We construct a ``navigation-manipulation'' task suite comprising 90 canonical task scenarios that range from localized manipulation to long-horizon composite navigation. Within this suite, a robot must perform discrete, grid-based navigation toward target waypoints and autonomously regulate its orientation for object alignment, while explicitly predicting pixel-wise bounding boxes (BBoxes) to facilitate subsequent manipulation. To enable effective decision-making in long-horizon tasks, RMMBench provides the complete textual history of past actions and environmental feedback, while restricting the visual context to a sliding window of recent observations to prevent context explosion. Through this operational interface, we establish a fine-grained evaluation protocol using 6 categories of process-oriented metrics to systematically analyze performance across these composite tasks.

The composite tasks in our benchmark are designed to evaluate embodied models in sequential navigation-manipulation coordination and long-horizon multi-goal planning.
Unlike existing benchmarks that rely on external navigation modules or predefined interaction triggers, we require models to autonomously perform navigation, spatial alignment, and manipulation based solely on continuous visual observations and environmental feedback.
This setting imposes strict requirements on low-level spatial grounding, temporal action consistency, and dynamic task scheduling in complex environments.
By explicitly modeling continuous execution processes and navigation-manipulation transitions, RMMBench enables fine-grained diagnosis of grounded spatial reasoning and low-level embodied decision-making beyond conventional task success metrics.

The main contributions of this work are summarized as follows:
\begin{itemize}
    \item We propose RMMBench, a comprehensive benchmark that directly assesses VLM-based methods for robotic navigation and manipulation from high-level language understanding and pixel-wise visual grounding to low-level action execution.
    \item We establish a fine-grained, multidimensional evaluation protocol to systematically decouple high-level planning failures from low-level execution bottlenecks, enabling a more precise diagnostic analysis.
    \item The evaluation results on RMMBench, which provides a ``navigation-manipulation'' task hierarchy comprising 90 canonical task scenarios, reveal that the spatial localization under complex visual conditions remains challenging for today's leading VLMs.
\end{itemize}

\begin{table}[t]
  \centering
  \caption{Comparison of benchmarks for embodied intelligence.
  Mobile Manip.: mobile manipulation tasks;
  Sem. Instruct.: rich semantic instructions;
  Vis. Ground.: native visual grounding via BBoxes;
  Act. Perception: active field-of-view recovery;
  Multi-View: multi-view integration;
  Hist. Obs.: historical visual observations;
  Hist. Context: historical decision context.}
  \label{tab:benchmark_comparison}
  \renewcommand{\arraystretch}{1.1}
  \resizebox{\textwidth}{!}{
  \begin{tabular}{l|ccccccc}
    \toprule
    Benchmark & Mobile Manip. & Sem. Instruct. & Vis. Ground. & Act. Perception & Multi-View & Hist. Obs. & Hist. Context \\
    \midrule
    VLABench~\cite{zhang2025vlabench}      & $\times$     & $\checkmark$ & $\times$     & $\times$     & $\checkmark$ & $\times$     & $\checkmark$ \\
    EMMOE~\cite{li2025emmoe}         & $\checkmark$ & $\checkmark$ & $\times$     & $\times$     & $\checkmark$ & $\times$     & $\checkmark$ \\
    VLMBench~\cite{zheng2022vlmbench}      & $\times$     & $\checkmark$ & $\times$     & $\times$     & $\checkmark$ & $\times$     & $\times$ \\
    EmbodiedBench~\cite{yang2025embodiedbench} & $\checkmark$ & $\checkmark$ & $\times$     & $\times$     & $\times$     & $\checkmark$ & $\checkmark$ \\
    RoboCasa~\cite{nasiriany2024robocasa}      & $\checkmark$ & $\checkmark$ & $\times$     & $\times$     & $\checkmark$ & $\times$     & $\times$ \\
    LIBERO~\cite{liu2023libero}        & $\times$     & $\times$     & $\times$     & $\times$     & $\checkmark$ & $\checkmark$ & $\times$ \\
    GenManip~\cite{gao2025genmanip}      & $\times$     & $\checkmark$ & $\times$     & $\times$     & $\times$     & $\times$     & $\checkmark$ \\
    LongBench~\cite{chen2026longbench}     & $\times$     & $\checkmark$ & $\times$     & $\times$     & $\checkmark$ & $\checkmark$ & $\times$ \\
    \textbf{RMMBench (Ours)} & $\checkmark$ & $\checkmark$ & $\checkmark$ & $\checkmark$ & $\checkmark$ & $\checkmark$ & $\checkmark$ \\
    \bottomrule
  \end{tabular}
  }
  \vspace{-2pt}
\end{table}

\section{Related Work}

\textbf{Embodied Manipulation Benchmarks.} The development of benchmarks and methodologies for embodied manipulation has transitioned from discrete symbolic planning to semantic reasoning driven by VLMs. Early long-horizon platforms \cite{dasari2019robonet,li2023behavior,heo2025furniturebench} focused primarily on the construction of low-level skill chains, yet their task spaces were strictly limited to stationary tabletop or cabinet-level operations. Although some room-scale environments \cite{li2023behavior,shridhar2020alfred} extended the diversity of scenarios, their underlying physical interactions often relied on discrete symbolic interfaces.

\textbf{Modular Planning Paradigms.} Embodied foundation models have diverged into end-to-end policies \cite{brohan2022rt,kim2024openvla,black2024pi0} and modular planning frameworks \cite{openai2024gpt,huang2024copa}. Nevertheless, state-of-the-art modular workflows \cite{duan2024manipulate} suffer from two key disconnections: one in perception and the other in interaction. On the perception side, they suffer from ``perceptual fragmentation'' by relying heavily on external, separate vision modules (e.g., SAM~\cite{ravi2025sam}, Grounding DINO~\cite{liu2024grounding}, or Set-of-Mark prompting \cite{jiang2023vima,yang2023set,shridhar2023perceiver}) to heuristically extract BBoxes. Feeding these as symbolic labels to the foundation model segregates semantic reasoning from native geometric spatial awareness. On the interaction side, existing platforms mostly lack deep closed-loop testing for the ``perception-execution disconnect''. When a VLM generates a faulty plan or encounters an unreachable state, prominent zero-shot evaluation frameworks \cite{huang2024copa, duan2024manipulate} typically terminate the current episode or rely on open-loop implicit corrections. To address these limitations, our work constructs a spatially extended ``navigation-manipulation'' composite task hierarchy, requiring models to coordinate navigation between task locations while maintaining native visual grounding during dynamic physical trial-and-error.

\textbf{Procedural Evaluation Metrics.} Although recent evaluation environments \cite{zhang2025vlabench,li2025emmoe} attempted to introduce diversified evaluation criteria, current frameworks still tend to favor macro-level, outcome-oriented metrics (such as final task success rates or coarse sub-task completion scores~\cite{jiang2023vima,fu2024mobile,goyal2024rvt,team2024octo,driess2023palm}) when quantifying subtle aspects such as redundant actions, wandering trajectories, or decision deadlocks. Such an evaluation bias easily obscures the procedural redundancy of agents during long-term execution. To bridge this gap, we propose a fine-grained diagnostic framework that achieves precise quantification of dynamic trial-and-error redundancy and subtle cognitive flaws in large models.

\section{RMMBench}

RMMBench requires VLM to execute continuous-space, long-horizon mobile manipulation tasks based on language instructions and raw visual feedback. Evaluating this capability requires a tight integration of sequential planning, spatial grounding, and closed-loop visual verification. In this section, we present the benchmark design: Sec.~\ref{subsec:Data} defines our multi-paradigm task hierarchy and atomic skill libraries; Sec.~\ref{subsec:Pipeline} elaborates the comprehensive evaluation pipeline across navigation and manipulation boundaries; and Sec.~\ref{subsec:Metrics} establishes the comprehensive suite for fine-grained performance evaluation.

\subsection{Task Hierarchy and Atomic Skill Libraries}
\label{subsec:Data}

To evaluate the embodied perception, semantic reasoning, and zero-shot execution capabilities of models, we construct a randomized data synthesis pipeline within a unified MuJoCo \cite{todorov2012mujoco} physics simulation environment supported by the dm\_control \cite{tunyasuvunakool2020dm_control} framework, leveraging RoboCasa \cite{nasiriany2024robocasa} and Objaverse \cite{deitke2023objaverse} assets. To improve physical plausibility, we apply geometric filtering to the raw assets and exclude objects whose scale makes them unsuitable for grasping. For composite navigation-manipulation tasks, the framework introduces diverse scenes from RoboCasa \cite{nasiriany2024robocasa} across multiple functional layouts and aesthetic styles. To evaluate policy robustness under distribution shifts, each scenario dynamically randomizes target object categories, initial coordinates, and scene material textures during initialization.

RMMBench comprises 90 canonical task scenarios structured across two primary evaluation paradigms to comprehensively stress-test VLMs. The first paradigm encompasses 60 standalone manipulation scenarios (39 short-horizon and 21 long-horizon tasks), spanning from localized structured executions with deterministic routines to open-ended interactions driven by high-level guidance. Since contemporary VLMs already possess rich world knowledge, we embed commonsense and semantic considerations into the natural-language instructions of both short- and long-horizon tasks rather than curating them as a separate category, focusing evaluation on the more challenging long-horizon tasks. Complementing these localized tasks, the second paradigm introduces 30 composite navigation-manipulation scenarios that demand cross-space skill chaining across multiple functional layouts. Manipulation task examples are illustrated in \Cref{fig:manip_visual}, and composite navigation-manipulation task examples in \Cref{fig:navigation1}, both in the supplementary material.

The core atomic skill library exposes discrete primitives including pick, place, pull, lift, push, and open\_door/close\_door, open\_sink, and rotate\_knob, alongside active ego-perception via wrist\_observe. When a VLM selects a skill, the DSL~\cite{nordmann2014survey,sutherland2019robolang} interface routes the command to a unified ``RRT--IK--low-level controller tracking'' execution pipeline: the VLM is responsible only for high-level decisions, including skill selection (with parameters), BBox prediction, and grasp candidate selection, while the backend handles RRT planning, IK solving, and low-level control. Skills such as lift and pull use preset default movement distances, and the VLM can increase the end-effector movement distance via re-execution.

\subsection{Evaluation Pipeline}
\label{subsec:Pipeline}

\begin{figure}[t]
\centering
\includegraphics[width=1\textwidth]{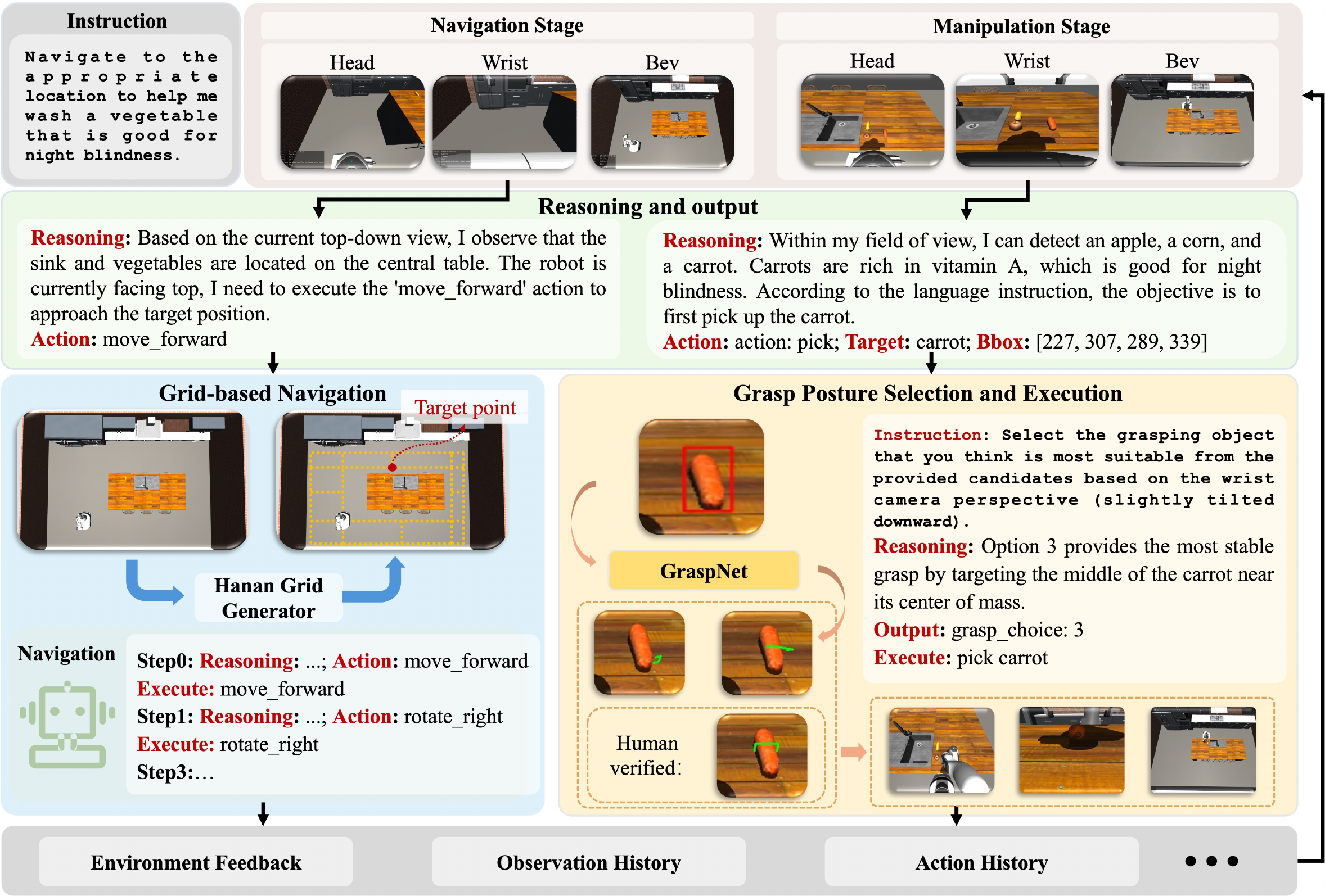}
\vspace{-3pt}
\caption{\textbf{Overview of the RMMBench evaluation pipeline.} It ingests head and wrist camera views, complemented by an optional bird's-eye-view (BEV) modality. Navigation is executed over a Hanan grid~\cite{hanan1966steiner} dynamically bounded by the workspace geometry. For manipulation tasks, the pipeline overlays algorithmic candidates from GraspNet~\cite{fang2020graspnet} and a human-verified baseline fallback onto the wrist view for the VLM to select. Decision-making is conditioned on the historical context.}
\label{fig:framework}
\vspace{-10pt}
\end{figure}

As illustrated in Figure \ref{fig:framework}, the RMMBench evaluation pipeline assesses the VLM’s multi-stage decision process through step-by-step rollouts. The framework provides the VLM with an interleaved context of text histories and visual streams, translating its macro-instructions into executable atomic skills, and conditionally routing explicit environmental feedback only when execution constraints are violated (e.g., collisions or out-of-reach errors).

\textbf{Grid-based Navigation.} The framework implements an adaptive movement specification over a discrete Hanan grid. When the VLM issues a forward command, the framework dynamically calculates the executed step size $d_{\text{step}}$ based on the maximum obstacle-free distance to the target grid point $d_{\text{max}}$ and a distance threshold $\tau$:

\begin{equation}
d_{\text{step}} =
\begin{cases}
d_{\text{init}}, & \text{if the grid point is undetected} \\
d_{\text{max}}, & \text{if } d_{\text{max}} \le \tau \\
\rho \cdot d_{\text{max}}, & \text{if } d_{\text{max}} > \tau
\end{cases}
\end{equation}
where $d_{\text{init}}$ is the baseline step size, and $\rho \in (0, 1)$ is the attenuation factor. This mechanism allows the robot to approach targets incrementally and stop precisely at the designated stay points once within $\tau$, ensuring arm-to-workstation alignment. If obstructed ($d_{\text{max}} \le 0$), the simulator blocks execution. This constraint design bypasses low-level path solvers to purely evaluate the spatial reasoning of the VLM.

\textbf{Tabletop Manipulation.} During manipulation tasks, the VLM uses head and wrist cameras to predict 2D BBoxes for targets and containers, and localization accuracy is evaluated using the intersection-over-union (IoU) between the predicted BBoxes and the corresponding ground-truth masks. For grasp execution, the options projected onto the wrist view comprise a hybrid candidate pool containing both dynamically generated GraspNet predictions and expert-verified ground-truth grasp annotations associated with each object asset. The latter serves as an environmental baseline fallback to guarantee candidate availability across complex scenes. Specifically, the target pose for pick is provided by the 6-DoF grasp pose generated by GraspNet, whereas the placement coordinates for place are determined by the VLM-predicted BBox; grasp points are bound to object assets and mapped to world coordinates via transformation matrices following the object's actual pose. Crucially, all pose candidates are anonymized and uniformly presented to the VLM. In tasks with confined spaces or specific directional constraints (e.g., horizontal or vertical approach requirements), GraspNet introduces necessary geometric directional diversity. The VLM must then autonomously perform zero-shot spatial reasoning to select the optimal option for low-level point-to-point path solvers.

\textbf{Action History and Error Feedback.} To support evaluations of long-horizon tasks, RMMBench maintains the full textual history of past actions while restricting the visual context to a sliding window of recent frames. The simulator enforces strict physical reachability and kinematic constraints without applying any implicit state corrections (e.g., automatic object teleportation or base repositioning). In most cases, the environment relies entirely on the subsequent sensory update, and the VLM must autonomously verify the execution outcome, such as detecting an empty gripper or an undisplaced target, solely through its visual observation stream. The only exception is text-based feedback during pick and place execution, which is provided under two conditions: (1) the target lies outside the reachable workspace; or (2) the predicted BBox contains no valid point cloud. All other failures (e.g., collisions, slippage, anomalous object displacement) produce no textual hints and expose no internal simulator state to the model. This design compensates for the model's lack of direct access to the robot’s kinematic constraints (e.g., arm length and reachable range), preventing the action budget from being exhausted on physically unreachable, ineffective operations. This non-oracle feedback mechanism rigorously tests whether the model can detect execution anomalies from raw vision and dynamically switch back to navigation or inspection primitives for repositioning and replanning.

\subsection{Evaluation Metrics}
\label{subsec:Metrics}

To provide a rigorous, non-binary evaluation of VLM performance in long-horizon composite tasks, RMMBench provides a quantitative evaluation suite across three core dimensions: task execution progress, visual grounding precision, and error-correction behavior. Specifically, task progression and baseline efficiency are decomposed into execution efficiency ($EE$), two completion rates, and a navigation progress measure. $EE$ penalizes redundant actions against the expert reference sequence length, and is computed only over successful trajectories. The manipulation completion rate ($CR_m$) measures the proportion of satisfied atomic sub-conditions. The navigation completion rate ($CR_n$) monitors workspace radius and orientation alignment. The navigation progress ($NP$) tracks spatial transit efficiency against the initial expert distance. To further evaluate high-level semantic alignment, the expert action recall rate ($RR_{\text{act}}$)~\cite{zhang2025vlabench} is formulated through the longest common subsequence (LCS) relative to the length of the expert reference sequence:
\begin{equation}
RR_{\text{act}} = \frac{\text{Len}(\text{LCS}(Seq_{\text{vlm}}, Seq_{\text{expert}}))}{\text{Len}(Seq_{\text{expert}})}
\end{equation}
where a single-step action constitutes a successful recall if both the skill primitive and the target entity category match the expert reference. For flexibly ordered tasks, an action is counted as recalled if it matches any valid permutation of the expert sequence that satisfies the underlying dependency constraints. The expert reference sequences are manually constructed, individually verified, and minimized.

To evaluate the spatial localization accuracy of a model across multi-view perception streams, RMMBench incorporates the visual grounding score ($AS_{\text{bbox}}$). It pools all 2D BBoxes predicted by the model across various execution stages, evaluating the alignment resolution via IoU against the ground-truth geometric mask of the target:
\begin{equation}
AS_{\text{bbox}} = \frac{1}{K} \sum_{k=1}^{K} \text{IoU} (\text{BBox}_k, \text{Mask}_k)
\end{equation}
where $K$ denotes the total number of visual grounding steps executed throughout the episode. Malformed or unparsable BBox outputs are assigned an IoU of 0.

Finally, to diagnose policy loops and failure recovery patterns under environmental negative feedback, RMMBench introduces the reflection score ($RS$), which aggregates all repeated-action loops within an episode. A loop is defined as the action sequence starting from the first grasp failure on a target until the model turns to manipulate another target; the sequence counts toward $RS$ only when the target is manipulated repeatedly. For an episode containing $M \geq 1$ loops of lengths $n_1, \dots, n_M$:

\begin{equation}
RS = \frac{1}{M} \sum_{i=1}^{M} \gamma_i \cdot e^{-\lambda n_i}
\end{equation}
where $\lambda > 0$ is a sensitivity hyperparameter controlling the decay rate over repeated ineffective actions, and $\gamma_i$ is a loop-level compliance coefficient that takes the value $1.0$ or $\alpha \in (0,1)$ depending on the ultimate success of the condition associated with the target of loop $i$, thereby enforcing a strict performance ceiling based on the ultimate outcome of the associated condition. To balance trajectory-level adaptability against ultimate task accomplishment, $RS$ incorporates all execution trials into the evaluation while structurally dampening failed attempts with an outcome-dependent weight. $RS$ is computed only over episodes containing at least one repeated-action loop.
\begin{equation}
\gamma =
\begin{cases}
1.0, & \text{if the associated condition is ultimately satisfied} \\
\alpha, & \text{otherwise}
\end{cases}
\end{equation}
where $\alpha \in (0, 1)$ serves as a baseline failure discount factor. Intuitively, this scoring rule encodes two premises: repeated ineffective actions on the same target reflect insufficient self-correction, and are therefore penalized by an exponential decay in loop length; moreover, the final outcome imposes a hard ceiling, as a loop whose associated condition is ultimately unmet can never score above $\alpha$. Importantly, $\alpha$ keeps the failure penalty tunable rather than absolute: a larger $\alpha$ retains partial credit for persistent attempts that ultimately fail, preventing genuine deadlock-recovery exploration from being crushed to zero, while, for loops of equal length, a failed loop always scores strictly below a successful one.

\begin{table*}[t]
\centering
\caption{Evaluation results across two manipulation task categories: short-horizon (39 scenarios) and long-horizon (21 scenarios) tasks, reported over 3 evaluation episodes per task.}
\vspace{2pt}
\label{tab:combined_manipulation_tasks_direct_metrics}
\renewcommand{\arraystretch}{1.05}
\resizebox{\textwidth}{!}{
\begin{tabular}{l|cccccc|cccccc}
\toprule
 & \multicolumn{6}{c|}{\textbf{Short-horizon Tasks}} & \multicolumn{6}{c}{\textbf{Long-Horizon Tasks}} \\
\cmidrule(lr){2-7} \cmidrule(lr){8-13}
\textbf{VLM Backend} & \textbf{SR} & $\text{CR}_m$ & \textbf{EE} & $\mathbf{RR}_{\text{act}}$ & $\mathbf{AS}_{\text{bbox}}$ & {$\mathbf{RS}$} & \textbf{SR} & $\text{CR}_m$ & \textbf{EE} & $\mathbf{RR}_{\text{act}}$ & $\mathbf{AS}_{\text{bbox}}$ & {$\mathbf{RS}$} \\
\midrule
{Gemini-3-Pro} & 58.12 & 60.29 & \textbf{78.94} & 73.50 & 49.05 & 51.37 & 42.86 & 59.70 & 71.48 & 63.89 & 32.72 & 35.99 \\
{Gemini-3-Flash} & 58.97 & 60.29 & 77.36 & 74.02 & 47.09 & 49.32 & 38.10 & 56.72 & 68.62 & 60.71 & 30.76 & 33.83 \\
{Qwen-3.5-Plus} & 41.03 & 41.18 & 54.74 & 50.94 & 18.90 & 12.58 & 14.29 & 26.87 & 48.00 & 28.57 & 14.52 & 11.03 \\
{Doubao-SEED-2.0-Pro} & 56.41 & 57.35 & 66.67 & 68.03 & 32.93 & 49.73 & 25.40 & 43.28 & 73.68 & 45.24 & 22.33 & 24.70 \\
{Gemini-2.5-Pro} & 56.41 & 57.35 & 72.80 & 68.03 & 18.92 & 41.11 & 33.33 & 47.76 & \textbf{77.14} & 51.19 & 26.79 & 41.25 \\
{Kimi-K3} & 41.88 & 42.65 & 71.29 & 61.03 & 29.07 & \textbf{55.74} & 17.46 & 29.85 & 57.87 & 31.15 & 19.30 & 39.06 \\
{DeepSeek-V4.1-Flash} & 22.22 & 23.53 & 48.33 & 50.26 & 10.72 & 45.68 & 3.17 & 8.96 & 31.73 & 24.21 & 3.84 & 26.71 \\
{GPT-6-Astra} & \textbf{88.89} & \textbf{90.20} & 78.45 & \textbf{94.53} & \textbf{52.08} & 55.63 & \textbf{58.73} & \textbf{74.63} & 74.17 & \textbf{79.37} & \textbf{56.26} & \textbf{65.76} \\
\bottomrule
\end{tabular}
}
\vspace{-12pt}
\end{table*}

\section{Experiments}
Our experimental design focuses on analyzing VLM behaviors across three core operational dimensions: (1) task progression on manipulation tasks of increasing complexity; (2) decentralized navigation capability relying solely on first-person active vision without global spatial priors; and (3) the stability and degradation patterns of 2D spatial grounding under long-horizon execution.

\subsection{Experimental Setups and Overall Performance}
We evaluate eight representative VLMs: Gemini-2.5-Pro~\cite{comanici2025gemini}, Gemini-3-Pro-preview, Gemini-3-Flash-preview, Qwen-3.5-Plus, Doubao-SEED-2.0-Pro-260215, GPT-6-Astra, Kimi-K3, and DeepSeek-V4.1-Flash. Tests are conducted in a unified MuJoCo simulation framework comprising 90 canonical task scenarios: 60 pure manipulation tasks and 30 composite navigation-manipulation tasks. Domain randomization is applied across these scenarios to support rigorous quantitative evaluation. Each task instance is evaluated over 3 independent episodes to guarantee statistical reliability. Task configurations are pre-generated as fixed JSON files, and all models load identical configurations to ensure fairness and reproducibility.

Experimental results demonstrate a clear decline in models' performance as task horizon and complexity increase. While most models exhibit a certain degree of generalization in short-horizon, single-interaction manipulation tasks, their success rates decrease when managing long-horizon tabletop tasks that demand sequential coordination. The performance drops significantly when transitioning further to composite navigation-manipulation tasks that require extended skill sequences. For these composite long-horizon scenarios, excluding the strongest model (GPT-6-Astra), the task success rates ($SR$) of the tested models fall to a low range of 3.3\%--21.1\% without BEV and 7.8\%--30.0\% with BEV, exposing severe limitations in maintaining sustained execution stability.

\textbf{Task Progression and Planning Performance.} We analyze the task completion and planning performance of a model across short- and long-horizon tasks, respectively. Table~\ref{tab:combined_manipulation_tasks_direct_metrics} shows that all models still translate natural-language requirements into executable action plans with moderate-to-high $RR_{\text{act}}$, indicating that contemporary VLMs already possess rich affordance knowledge.

\begin{figure}[t]
\centering
\includegraphics[width=1\textwidth]{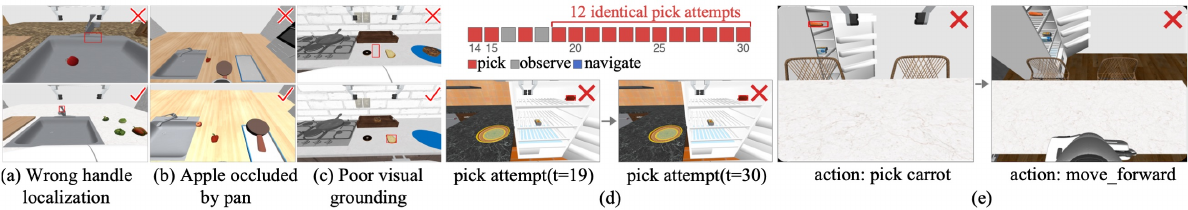}
\vspace{-15pt}
\caption{\textbf{Typical failure cases in task execution.} (a--c) Manipulation failures (top: failed executions; bottom: correct executions) involving handle mis-localization, target occlusion, and inaccurate visual grounding. (d) Repeated picking of targets beyond the operational range. (e) Outputting pick and then executing move\_forward despite the presence of obstacles.}
\label{fig:failure_cases}
\vspace{-10pt}

\end{figure}

However, for long-horizon manipulation tasks that require sequential action coordination, the overall success rate decreases. This degradation is primarily driven by localized execution failures. As shown in Figure~\ref{fig:failure_cases} (a--c), these localized failures mainly manifest in three modes: localization deviations of manipulation targets (e.g., the faucet handle), decision-making errors when targets are occluded, and insufficient BBox grounding accuracy. Further case-level analyses are provided in \Cref{app:qualitative}, and representative rollouts of GPT-6-Astra are shown in \Cref{app:astra_examples} of the supplementary material. Specifically, when BBox accuracy drops and triggers an interaction failure, the model repeats the exact same failed action primitive on the target entity instead of using inspection primitives such as wrist camera observations to regain a clear field of view and replan the execution. This operational pattern decreases $RS$, exposing a critical limitation in dynamic closed-loop error correction and feedback-driven strategy adaptation during extended physical interactions.

\begin{table*}[t]
\centering
\caption{Evaluation results of composite navigation tasks with BEV mapping.}
\vspace{2pt}
\label{tab:bev_composite_nav_results}
\renewcommand{\arraystretch}{1}
\resizebox{\textwidth}{!}{
\begin{tabular}{l|cccccccc}
\toprule
\textbf{VLM Backend} & \textbf{SR (\%)} & $\mathbf{CR_n}$ (\%) & {$\mathbf{NP}$} & $\mathbf{CR_m}$ (\%) & \textbf{EE} & $\mathbf{RR_{\text{act}} }$ & $\mathbf{AS_{\text{bbox}} }$ & {$\mathbf{RS}$} \\
\midrule
{Gemini-3-Pro}& 25.56 & 34.44 & 45.50 & 37.21 & 48.33 & 38.95 & 34.34 & 38.53 \\
{Gemini-3-Flash}& 30.00 & 34.07 & 45.21 & 35.66 & 46.42 & 34.39 & 32.49 & 36.50 \\
{Qwen-3.5-Plus} & 8.89 & 10.00 & 16.31 & 11.63 & 39.63 & 9.82 & 15.12 & 11.32 \\
{Doubao-SEED-2.0-Pro} & 21.11 & 25.56 & 36.19 & 28.68 & 45.35 & 28.42 & 20.75 & 32.32 \\
{Gemini-2.5-Pro} & 17.78 & 20.74 & 30.43 & 25.58 & 46.32 & 24.56 & 16.08 & 26.31 \\
{Kimi-K3} & 12.22 & 13.33 & 21.48 & 14.73 & 39.70 & 12.98 & 15.99 & 30.66 \\
{DeepSeek-V4.1-Flash} & 7.78 & 5.93 & 9.80 & 6.20 & 38.87 & 5.61 & 9.65 & 20.56 \\
{GPT-6-Astra} & 48.89 & 54.07 & 67.86 & 58.91 & 53.53 & 55.09 & 40.62 & 47.29 \\
\bottomrule
\end{tabular}
}
\vspace{-8pt}
\end{table*}

\begin{table*}[t]
\centering
\caption{Evaluation results of composite navigation tasks without BEV mapping.}
\vspace{2pt}
\label{tab:no_bev_composite_nav_results}
\renewcommand{\arraystretch}{1}
\resizebox{\textwidth}{!}{
\begin{tabular}{l|cccccccc}
\toprule
\textbf{VLM Backend} & \textbf{SR (\%)} & $\mathbf{CR_n}$ (\%) & {$\mathbf{NP}$} & $\mathbf{CR_m}$ (\%) & \textbf{EE} & $\mathbf{RR_{\text{act}} }$ & $\mathbf{AS_{\text{bbox}} }$ & {$\mathbf{RS}$} \\
\midrule
{Gemini-3-Pro}& 17.78 & 22.59 & 34.88 & 24.81 & 46.94 & 23.86 & 29.43 & 31.85 \\
{Gemini-3-Flash}& 21.11 & 24.44 & 36.12 & 27.91 & 46.15 & 27.37 & 29.20 & 30.09 \\
{Qwen-3.5-Plus} & 5.56 & 7.78 & 14.32 & 10.08 & 38.87 & 8.07 & 13.23 & 10.06 \\
{Doubao-SEED-2.0-Pro} & 11.11 & 15.93 & 26.20 & 18.60 & 42.60 & 18.60 & 18.44 & 25.36 \\
{Gemini-2.5-Pro} & 10.00 & 14.81 & 25.52 & 17.83 & 43.51 & 17.19 & 13.81 & 20.56 \\
{Kimi-K3} & 6.67 & 6.67 & 11.53 & 7.75 & 39.92 & 6.67 & 12.79 & 23.41 \\
{DeepSeek-V4.1-Flash} & 3.33 & 5.19 & 8.81 & 6.20 & 37.64 & 5.26 & 8.58 & 15.53 \\
{GPT-6-Astra} & 40.00 & 47.04 & 63.35 & 55.04 & 52.61 & 51.23 & 36.46 & 40.61 \\
\bottomrule
\end{tabular}
}
\vspace{-5pt}
\end{table*}

\textbf{Active Vision and Egocentric Navigation.} We conduct an ablation study to compare navigation performance with and without BEV mapping. The results listed in Tables~\ref{tab:bev_composite_nav_results} and~\ref{tab:no_bev_composite_nav_results} reveal a critical bottleneck: stripping away global top-down priors leads to a severe degradation in both navigation success rates and pathfinding efficiency across all models. Under purely egocentric constraints with discrete movement and inspection action spaces, the model tends to rely heavily on rotation primitives to keep the target object within its local field of view rather than proactively exploring the environment via forward advancement, which drastically increases action redundancy.

Furthermore, due to the lack of global spatial references, a VLM tends to prematurely issue localized manipulation commands before the mobile base reaches the workstation interaction radius, as shown in Figure~\ref{fig:failure_cases}(d), some models persist in grasping targets that are visible yet beyond the operational range, repeatedly issuing identical pick commands instead of navigating closer, which is an indication of inadequate perception of their own operational boundaries. Similarly, as shown in Figure~\ref{fig:failure_cases}(e), models fail to leverage historical visual context to avoid previously observed obstacles, still outputting pick followed by move\_forward directly toward the obstacle. These failures expose the fragility of egocentric spatial reasoning and temporal visual memory in VLMs.

\textbf{Spatial Grounding Degradation in Long Sequences.} We probe the effect of context length on localization accuracy via a controlled experiment that fixes the visual input and varies only the number of history frames in the context (2--16), with full results in Table~\ref{tab:iou_detail} of the supplementary material. Localization accuracy varies in a clearly model-dependent manner: models such as Gemini-3-Flash remain largely stable, whereas only SEED-2.0 exhibits a monotonic degradation that emerges earlier and is exacerbated under the long-distance, occluded condition, indicating that context length alone is not the direct cause but interacts with scene complexity and model-dependent visual robustness.

\section{Conclusion}
\label{sec:conclusion}
We introduce RMMBench, a simulation benchmark for evaluating VLMs on grounded spatial perception and composite mobile manipulation, whose comprehensive evaluation suite decouples planning efficacy from localized execution failures to move beyond conventional outcome-oriented evaluation. Experimental results show that state-of-the-art VLMs, while possessing adequate affordance knowledge for short-horizon manipulation, exhibit weak egocentric pathfinding when detached from global top-down priors, with spatial localization robustness under long sequences being model-dependent—models sensitive to complex visual conditions degrade markedly as sequences accumulate. These findings reveal critical bottlenecks in modular VLM workflows and underscore the need for stronger temporal visual memory and dynamic error correction for robust embodied deployment.

\section{Limitations and Future Work}
The current scope of RMMBench is limited by the number of canonical task scenarios and the granularity of the atomic skill library. The navigation pipeline adopts a discretized grid abstraction, which isolates spatial reasoning but oversimplifies unstructured real-world terrain. The manipulation pipeline employs a frozen, off-the-shelf GraspNet for generalization, introducing localized grasping errors. Moreover, our controlled experiments show that grounding degradation with growing context length is model-dependent: only SEED-2.0 degrades persistently, exacerbated under long-distance and occlusion conditions. Future work will scale the task hierarchy, integrate head observation primitives, fine-tune perception-grasping modules, and further characterize the interplay between task horizon and grounding precision.

\acknowledgments{This work was supported in part by the National Natural Science Foundation of China under Grant U22A2057, in part by the Key R\&D Program of Shandong Province under Grant 2025CXGC010210, and in part by the Municipal-University Collaborative Development Project of Jinan under Grant JNSX2025002.}
\clearpage
\bibliography{example}

\clearpage
\appendix
\counterwithin{figure}{section}
\counterwithin{table}{section}
\counterwithin{equation}{section}

\begin{center}
    {\LARGE\bfseries Supplementary Materials for RMMBench}
\end{center}
\vspace{1em}

\section{Skill Library}

\subsection{Atomic Skills}
The library provides discrete primitives executed via an inverse kinematics (IK) solver. All manipulation skills are executed in a point-to-point manner, where low-level joint trajectory planning is resolved using a rapidly-exploring random tree (RRT) path planning solver. Parameterized skills primarily consist of pick $\langle$object$\rangle$ and place to $\langle$container$\rangle$. Non-parameterized manipulation skills include lift, pull, and push, which command the gripper to translate along fixed vertical or horizontal step sizes pre-configured by the system, while using the IK and RRT solvers to resolve the required 6-DoF end-effector trajectories. Sensory and state primitives include close\_gripper, open\_gripper, open\_door, close\_door, and observe, where observe resets the arm to a default canonical configuration to stabilize the visual field. Discrete navigation primitives regulate the mobile chassis coordinates in the environment layout and comprise move\_forward, rotate\_left, and rotate\_right. The orientation skills command the chassis to execute fixed-angle rotations about its vertical axis.

\begin{figure}[H]
\centering
\includegraphics[width=1\textwidth]{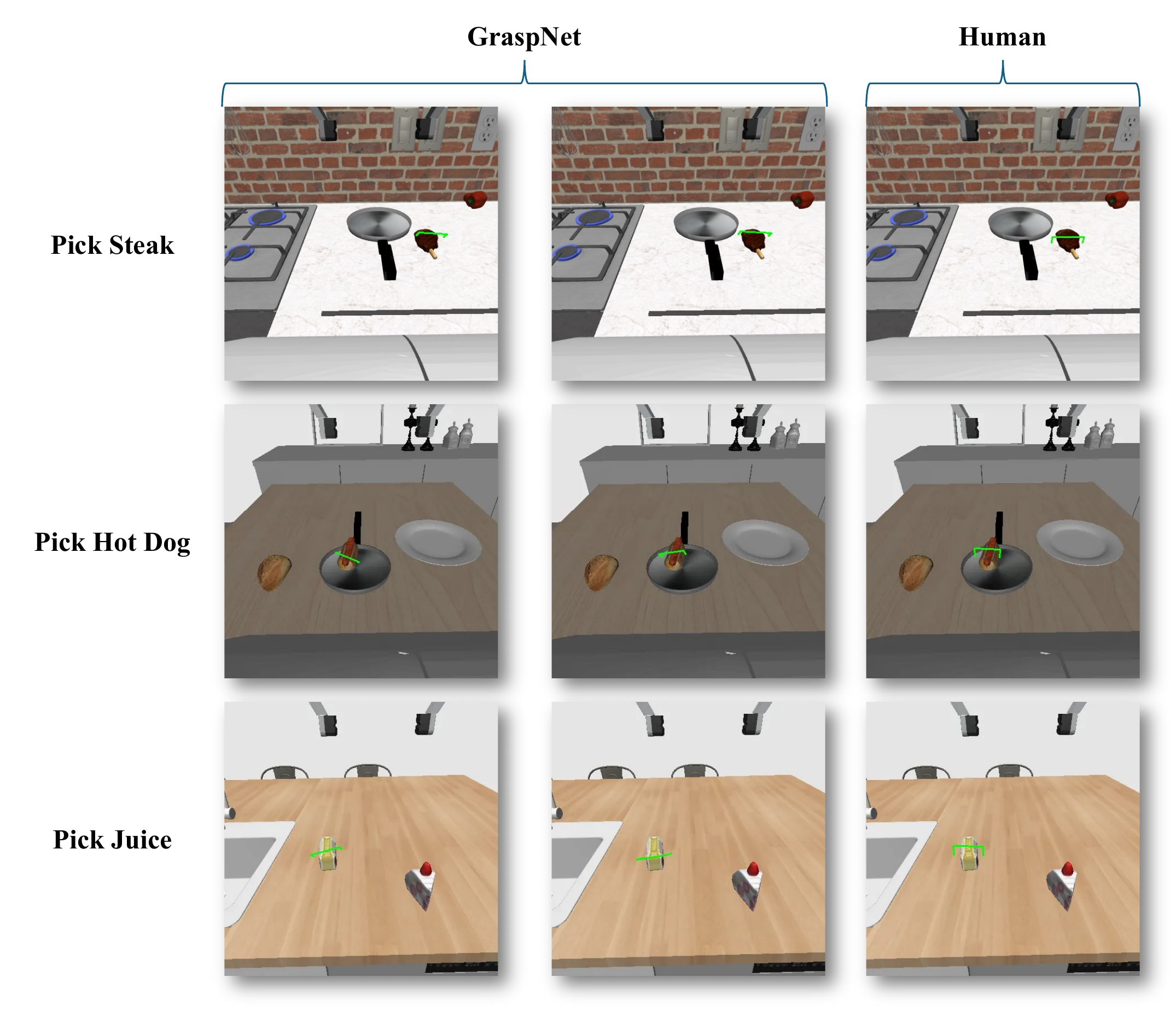}
\caption{Examples of grasp selection. Each selection round contains three grasp visualization images, where two are generated by GraspNet~\cite{fang2020graspnet} to prevent object orientation changes, and one is a manually defined fixed grasp pose.}
\label{fig:grasp}
\end{figure}

\subsection{Grasp Tuning and Filtering}
\textbf{Grasp Selection Pipeline}. The detailed prompt structure designed for this decision-making process is presented below. Grasp proposals generated by a GraspNet~\cite{fang2020graspnet} backend are presented as a
series of standalone images overlaying a visual template. Each candidate image isolates a single
grasp option, visually rendered as a green rectangular open-jaw gripper. These options are unlabeled regarding their generative source and are presented sequentially to the vision-language model (VLM) without sorting
prioritization. 

\begin{tcolorbox}[
    enhanced,
    breakable,                      
    boxrule=1.5pt,                  
    colframe=brown!80!black,        
    colback=blue!5!white,           
    colbacktitle=brown!70!black,    
    title={Grasp Selection Prompt for VLM}, 
    fonttitle=\large\bfseries\sffamily, 
    arc=6pt,                        
    sharp corners=south,            
]
\textbf{System Prompt:} \\
\small 
Role: You are an expert in robotic arm grasp planning. \\
\\
Task: pick \{target\_entity\} \\
\\
Rule: \\
- Select the grasping object that you think is most suitable from the provided candidates based on the wrist camera perspective (slightly tilted downward). \\
- Indicator: Green open-ended rectangular wireframes.The opening direction of the wireframe indicates the gripper's Approach Axis. \\
\\
Output Format Rules: \\
- Output the choice in the format: choice: 1/2/3/4... \\
- You should optionally output your reasoning process in the format: reasoning: \textless{}your step-by-step analysis\textgreater{} \\
\\
Examples: \\
- Case 1 (With reasoning): \\
choice: 1 \\
reasoning: your reasoning
\end{tcolorbox}

Specifically, the grasp candidates in our task include both autonomous proposals generated directly by GraspNet~\cite{fang2020graspnet} during the process without human verification, and human-selected grasp points which consist of GraspNet-generated options screened by humans as well as object geometric centers. The visualization of these grasp selection examples is shown in \Cref{fig:grasp}.

To isolate reachable poses for the autonomously generated category, the raw GraspNet proposals undergo a four-layer filter. First, 3D grasps are projected onto the 2D image plane, removing candidates outside the predicted object bounding box. Second, a model-free collision detector discards any pose with a clearance distance under 0.01 m or an approach distance within 0.05 m against the scene point cloud. Third, candidates are screened based on depth and position percentiles within the camera frame, keeping the closest 10\% of grasps while restricting the coordinates to the upper-middle regions. Fourth, the remaining valid candidates are ranked by their network score, preserving the top 4 highest-scoring configurations as actionable choices for the RRT-based execution.

\subsection{Placement Selection}
\begin{figure}[t]
\centering
\includegraphics[width=1\textwidth]{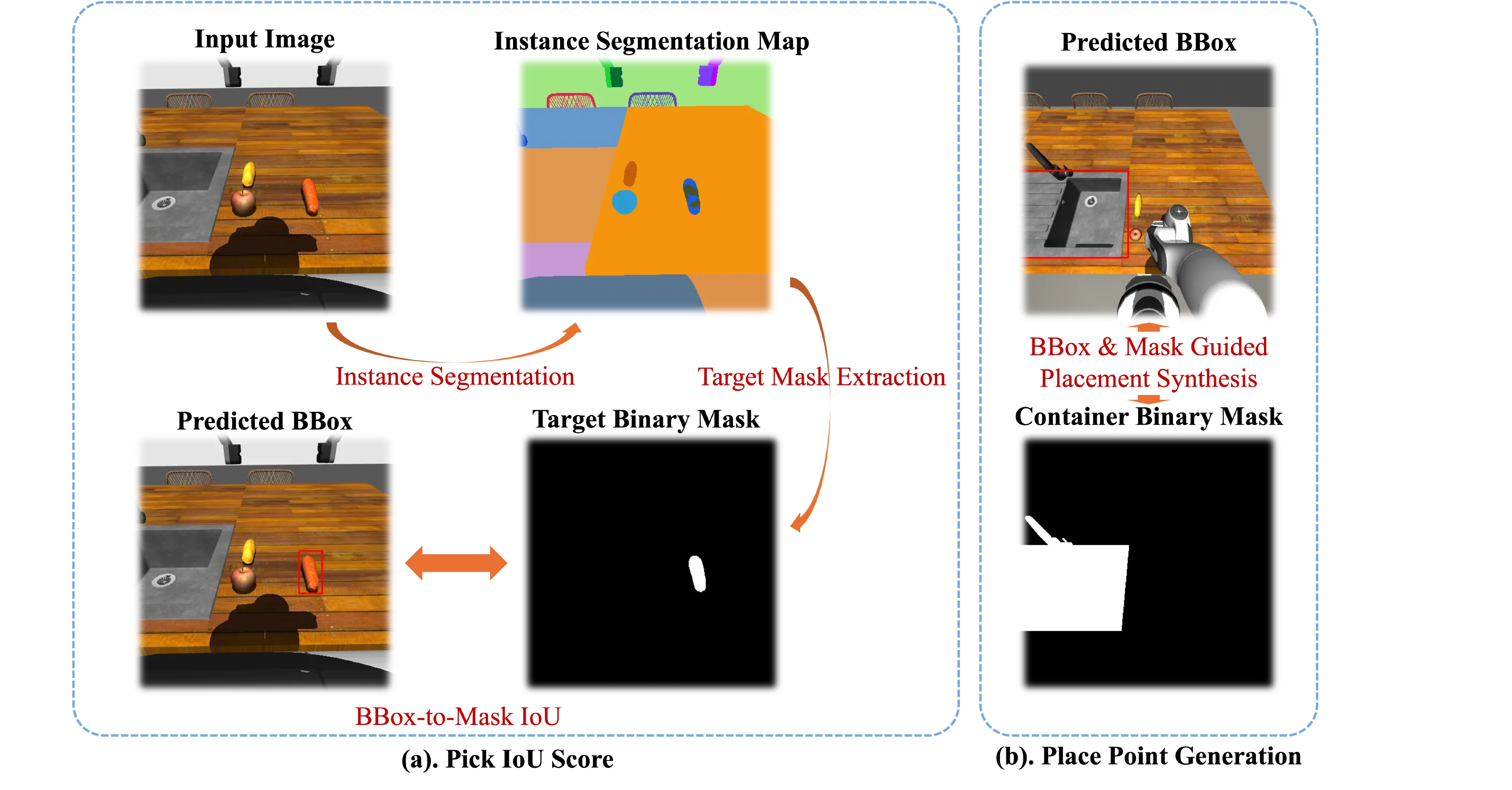}
\caption{(a) Pick IoU Score: Evaluates the VLM's grounding accuracy on target objects.
(b) Place Point Generation: Acquires the 3D placement coordinate via hierarchical sampling.}
\label{fig:pickplace}
\end{figure}

\textbf{Placement Pose Acquisition.} Terminal placement success is determined via axis-aligned bounding box inclusion testing, where the object 3D bounding box must be enclosed within the container volume. For flat surfaces, a vertical constraint of $\Delta z \le 1.5\,\text{cm}$ is enforced to prevent floating placement. The 3D target coordinate for the place to primitive is resolved across four sequential stages within the selected bounding box. First, Image Parsing captures the RGB, depth, and segmentation streams to locate the box center. Second, Container Mask Extraction cross-references the container name with the segmentation IDs to generate a local semantic mask. Third, Hierarchical Fallback Sampling selects the target landing pixel $(u, v)$ via a three-tier strategy: Strategy A morphologically erodes the container mask and selects pixels within the closest 10\% distance to the box center; Strategy B uses the original un-eroded mask if erosion empties the pixels; Strategy C isolates the furthest 30\% of depth pixels if no container mask is detected to avoid arm self-occlusion. Finally, Re-projection and Offset Bias converts the sampled 2D pixel and depth value into 3D world space using the inverse camera matrices, extracts the 3D median to filter outliers, and adds a vertical safety height for the RRT trajectory planner.The overall workflow of this coordinate acquisition process is illustrated in \Cref{fig:pickplace}(b).

\begin{figure}[t]
\centering
\includegraphics[width=1\textwidth]{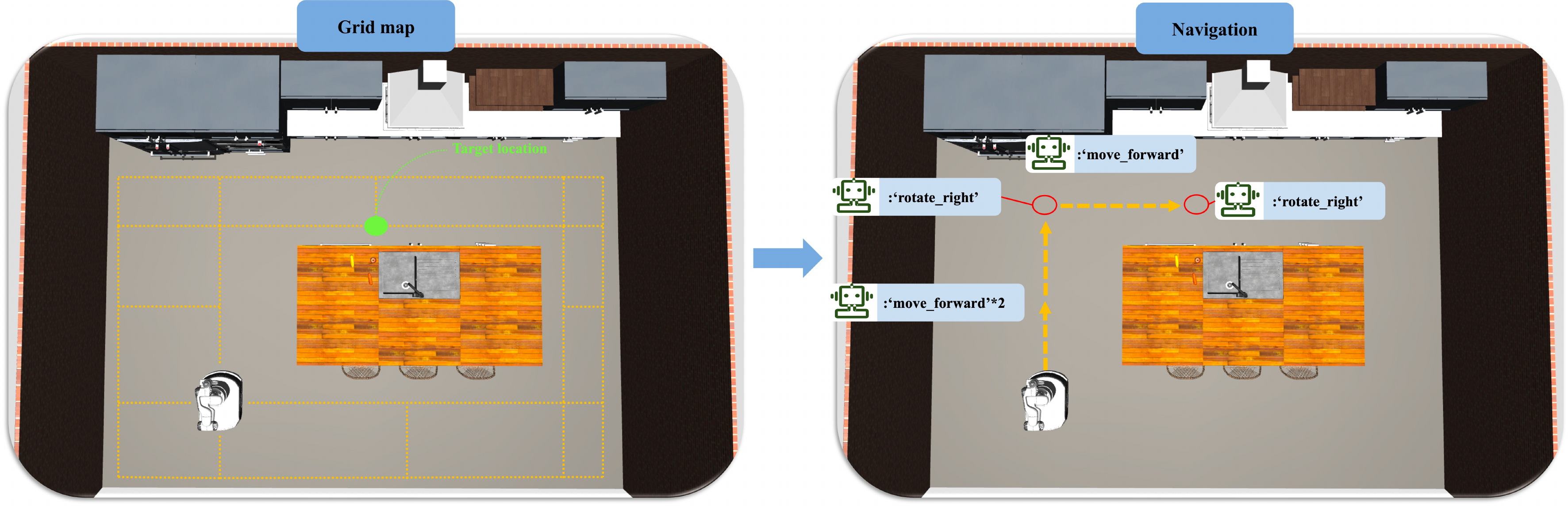}
\caption{An illustrative example of the Hanan-grid-based adaptive navigation mechanism, showcasing the distance-driven stepping control flow.}
\label{fig:hanangrid}
\end{figure}

\subsection{Hanan Adaptive Navigation}
\textbf{Distance Prediction.} The move\_forward skill discards rigid, fixed-step tracking to ensure safe and efficient motion near obstacles. Instead, it leverages a continuous-space adaptive stepping mechanism governed by a localized Hanan grid~\cite{hanan1966steiner} framework. Prior to navigation, the system constructs a target point pool containing the 2D world coordinates of the instruction targets and obstacles. To avoid zero-step self-locking, any target point within two robot radii of the chassis is filtered out using Manhattan distance. The system then casts a virtual ray from the robot heading direction into the MuJoCo~\cite{todorov2012mujoco} layout, referencing the room bounds and obstacle oriented bounding boxes to calculate the maximum obstacle-free distance $L_{\max}$ before encountering a boundary or a target point.

\textbf{Stepping Control.} Once $L_{\max}$ is resolved, the actual forward execution step size $l_{\text{step}}$ is calculated dynamically through a tri-conditioned control flow utilizing a security threshold $L_{\text{th}} = 1.2\,\text{m}$:
\begin{equation}
l_{\text{step}} =
\begin{cases}
l_{\text{default}}, & \text{if } L_{\max} = \infty \\
L_{\max}, & \text{if } L_{\max} < L_{\text{th}} \\
\frac{L_{\max}}{2}, & \text{if } L_{\max} \ge L_{\text{th}}
\end{cases}
\end{equation}
Under open-boundary scenarios where no obstacle or node is detected ($L_{\max} = \infty$), the model operates in a conservative exploratory mode, enforcing a short default step $l_{\text{default}} = 0.4\,\text{m}$. When the chassis approaches a workstation within close proximity ($L_{\max} < L_{\text{th}} $), the primitive commands a full $L_{\max}$ step to achieve a zero-error terminal docking alignment precisely at the target boundary. For far-horizon transits ($L_{\max} \ge L_{\text{th}} $), a half-distance binary reduction strategy is applied to mitigate deceleration latency and prevent high-inertia kinetic collisions within the physics engine. To execute this movement, the backend control loop runs a standard linear interpolation over the calculated displacement $l_{\text{step}}$ and directly streams the interpolated waypoints into the low-level chassis actuators frame-by-frame(with a visual demonstration of this Hanan-grid-based adaptive navigation shown in \Cref{fig:hanangrid}).

\section{Tasks and Environments}

\begin{figure}[t]
\centering
\includegraphics[width=1\textwidth]{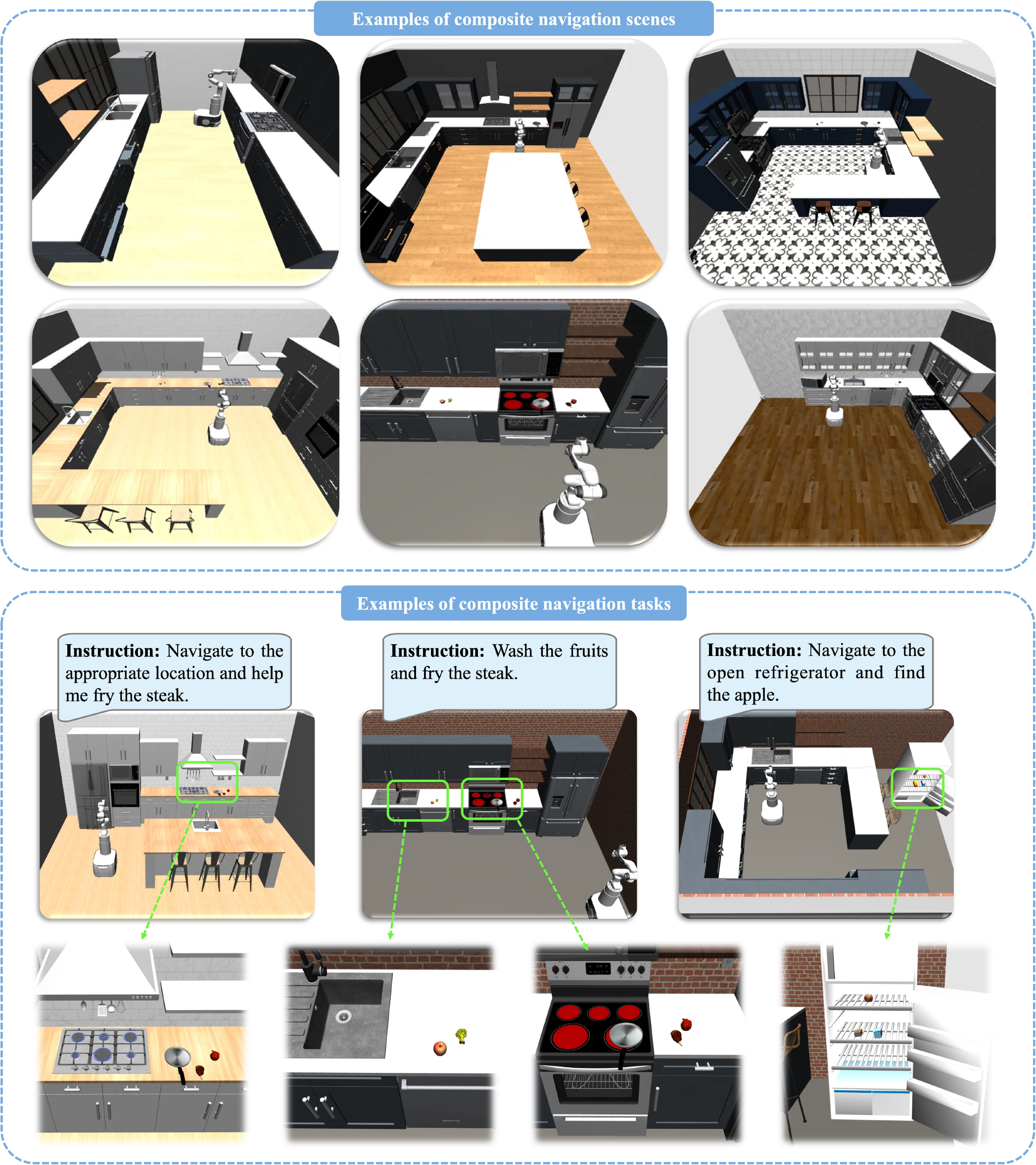}
\caption{Examples of composite navigation-manipulation tasks, showcasing diverse Robocasa-style functional layouts and instruction-guided task examples.}
\label{fig:navigation1}
\end{figure}

\subsection{Task Suite Construction}
The benchmark consists of 90 canonical scenarios whose natural-language instructions embed commonsense and semantic requirements, thereby evaluating long-horizon planning and embodied reasoning. These scenarios are split into two distinct categories: 60 tabletop manipulation tasks with stationary bases (39 short-horizon and 21 long-horizon tasks), and 30 composite navigation-manipulation tasks that require moving across functional layouts. The manipulation tasks focus on multi-step sorting, cooking procedures, and sequential goal planning. The composite tasks focus on coordinating the transitions between base locomotion and arm interaction, with various layout styles and instruction-guided task examples illustrated in \Cref{fig:navigation1}.

\subsection{Simulation Setup}
The evaluation architecture is built inside the MuJoCo~\cite{todorov2012mujoco} physics engine via the dm\_control~\cite{tunyasuvunakool2020dm_control} framework. The mobile manipulator agent is modeled on an Omron omnidirectional chassis integrated with a 7-DoF Franka Emika robotic arm and a parallel-jaw gripper. Visual streams from the torso-mounted head camera and the wrist-mounted camera are rendered at a uniform resolution of $480 \times 480$ pixels. The simulation operates at a baseline control timestep of 0.001 s, with all point-to-point arm trajectories fully managed by the low-level kinematic solvers.

\subsection{Domain Randomization}
To evaluate policy generalization under distribution shifts, the framework randomizes target object positions and environment styles upon initialization. Spatial object positioning utilizes a farthest first sampling algorithm over valid surface grids to prevent overlapping. For composite tasks, the initial mobile chassis coordinates remain fixed to evaluate navigation efficiency independently. Visual randomization dynamically alters the overall style of RoboCasa scenes and varies the workspace surface textures across different episodes.

\section{Evaluation Metrics}

\subsection{Task Progress}
To provide a non-binary quantification of embodied performance, RMMBench formalizes multiple continuous progress and tracking metrics. Execution efficiency ($EE$) penalizes redundant action outputs and is computed exclusively for successful trajectories as:
\begin{equation}
EE = \frac{\text{Len}(Seq_{\text{expert}})}{\text{Len}(Seq_{\text{vlm}})}
\end{equation}
where $EE \in (0, 1]$. Note that $EE$ is computed exclusively over successful trajectories: failed trajectories are excluded from both the computation and the averaging of this metric, preventing trajectory redundancy from being confounded with the ultimate task outcome. 

The manipulation completion rate ($CR_m$) measures structural sub-task completion and is defined as:
\begin{equation}
CR_m = \frac{N_{\text{satisfied}}}{\text{Len}(Seq_{\text{conditions}})}
\end{equation}
where $N_{\text{satisfied}}$ represents the cleared required sub-goal atomic conditions within an episode. 

\textbf{Built-in process conditions are implemented within the evaluation framework to track task completion.} Containment and removal tests verify whether a target entity is geometrically enclosed inside a designated container, or conversely, confirm it has been completely extracted and has reached a stable, at-rest equilibrium. State and contact audits check whether the robotic jaws maintain a valid friction-locked grasp on the target object, and monitor raw physical contact events between interacting entity meshes. Logical execution gates evaluate complex multi-step routines by tracking whether distinct sub-condition groups are satisfied either sequentially over time or via alternative parallel success states.

The navigation completion rate ($CR_n$) evaluates locomotion and alignment precision using a multi-tiered step function based on target zone entry and heading verification:
\begin{equation}
CR_n = 
\begin{cases} 
1.0, & \text{if the robot enters the target zone and achieves correct heading alignment} \\
0.5, & \text{if the robot enters the target zone but fails heading alignment} \\
0.0, & \text{otherwise}
\end{cases}
\end{equation}
The navigation status of each stop point retains the best state achieved throughout the episode: once the robot has arrived at a stop point, the recorded arrival and heading-alignment results are preserved even if the robot subsequently moves away.

Transit planning over layout structures is tracked via the navigation progress ($NP$), whose reference path is computed over the stop points prescribed by the task. The expert reference path is defined as the sum of the shortest path lengths from the robot's initial position through the stop points in the prescribed order:
\begin{equation}
D = d(p_{\text{init}}, s_1) + \sum_{j=1}^{N-1} d(s_j, s_{j+1})
\end{equation}
where $d(\cdot, \cdot)$ denotes the shortest path distance computed via Dijkstra's algorithm~\cite{dijkstra2022note} over the localized Hanan grid~\cite{hanan1966steiner}, and $s_1, \dots, s_N$ are the $N$ stop points prescribed by the task. Stop points are checked sequentially in the prescribed order: a stop point is regarded as passed once the robot has arrived at it at any point during the episode. Let $s_k$ denote the first stop point that has not been passed in this sequential check; the remaining path length is computed from the robot's final position through this stop point and all subsequent ones:
\begin{equation}
D_{\text{rem}} = d(p_{\text{final}}, s_k) + \sum_{j=k}^{N-1} d(s_j, s_{j+1})
\end{equation}
The navigation progress is defined as the proportion of the expert reference path that has been completed:
\begin{equation}
NP = \max \left( 0, \, \frac{D - D_{\text{rem}}}{D} \right)
\end{equation}
In the ideal case where the robot passes all stop points in order, $NP$ reaches its upper bound of 1.

The expert action recall rate ($RR_{\text{act}}$) is formulated in the main paper through the longest common subsequence relative to the length of the expert reference sequence; here we describe its implementation. A single-step action constitutes a successful recall only if both the skill primitive and the operating target are matched, where the operating target is identified by the actual object resolved from the visual grounding (predicted BBox), and correspondence between a facility (e.g., a sink or a stove) and its operating parts (e.g., a handle or a knob) is supported during matching. Observation actions are removed from both sequences prior to matching, participating in neither the matching nor the denominator.

Each task can be decomposed into a set of operation stages, each beginning with a grasp and accomplishing a single relocation or state change of the target object (e.g., \emph{picking the apple and placing it into the sink} and \emph{grasping the stove knob and turning it on} constitute two separate operation stages). For flexibly ordered tasks, the operation stages may be completed in arbitrary order; matching enumerates all stage permutations that satisfy the underlying dependency constraints (the temporal order within each stage is preserved), computes the recall for each permutation, and takes the maximum. For strictly ordered tasks, the reference sequence is matched directly. In composite navigation tasks, navigation skills are likewise removed from both sequences before matching, so that only the recall of manipulation skills is evaluated. 

\subsection{Grounding and Reflection}
The visual grounding score ($AS_{\text{bbox}}$) quantifies 2D localization accuracy across multi-view streams, with a representative example of the Pick IoU evaluation illustrated in \Cref{fig:pickplace}(a). All 2D BBoxes predicted by the model across various execution stages are pooled and evaluated via the intersection-over-union ($\text{IoU}$) against the ground-truth geometric mask of the target, following the formulation in the main paper:
\begin{equation}
AS_{\text{bbox}} = \frac{1}{K} \sum_{k=1}^{K} \text{IoU} (\text{BBox}_k, \text{Mask}_k)
\end{equation}
where $K$ denotes the total number of grounding steps. Every grounding step is counted in the denominator: malformed or unparsable BBox outputs are assigned an IoU of 0 rather than excluded from the statistics, thereby penalizing format-compliance failures and coordinate-estimation errors jointly. Since existing models all output the standard format, this treatment does not affect the reported results.

To diagnose policy loops and failure recovery behaviors under negative environmental feedback, the reflection score ($RS$) aggregates all repeated-action loops within an episode following the formulation in the main paper; here we present its implementation. An episode is decomposed into operation stages following the stage decomposition described above for $RR_{\text{act}}$, where repeated grasps of the same target are treated as within-stage retries. Only stages in which repeated manipulation occurs participate in the computation: stages involving a single grasp (regardless of whether the associated condition is ultimately satisfied) and targets that participate in no task condition are both excluded, preventing unrelated targets from dragging down the score. Episodes containing no repeated-action loop are likewise excluded from $RS$.

The loop length $n_i$ is counted from the first grasp failure on the corresponding target, excluding the failing step itself and observation actions. The compliance coefficient $\gamma_i$ is defined at the stage level and determined by the ultimate status of the task condition associated with the target of loop $i$: it takes the value 1.0 if the associated condition is ultimately satisfied, and otherwise the failure discount factor $\alpha \in (0,1)$. Throughout all experiments, we set $\lambda = 0.2$ and $\alpha = 0.5$.

\section{Prompt Design}
To effectively leverage the zero-shot reasoning capabilities of VLMs without fine-tuning, RMMBench implements a structured, state-conditioned prompting pipeline. Upon task initialization, the framework dispatches one of two distinct system prompt templates based on the underlying task paradigm. For localized interaction, models receive a dedicated manipulation template focusing on spatial grounding and end-effector coordinate selection. For long-horizon tasks, models receive a mobile navigation template (originally designed as composite\_navigation) that explicitly decouples execution into sequential phases, establishing a logical priority where navigation and alignment must be completed before manipulation skills are triggered. To analyze multi-modal flexibility, the mobile navigation template dynamically adapts its instructional field based on sensor configurations, supporting both standard dual-camera inputs (mobile\_navigation\_no\_bev) and three-view inputs augmented by a top-down bird's-eye view (BEV) modality (mobile\_navigation). The system prompts deployed for the manipulation and mobile manipulation tasks are structured as follows.


\begin{tcolorbox}[
    enhanced,
    breakable,                      
    boxrule=1.5pt,                  
    colframe=brown!80!black,        
    colback=blue!5!white,           
    colbacktitle=brown!70!black,    
    title={Manipulation System Prompt}, 
    fonttitle=\large\bfseries\sffamily, 
    arc=6pt,                        
    sharp corners=south,            
]
\textbf{System Prompt:} \\
\small 
You are a robotic arm with seven links, mounted on a white mobile base, situated in a given room environment. You need to complete tasks according to human instructions. \\
The system provides a set of available actions (Available\_Actions) with descriptions for each. At every step, you should choose one action from the available action set to execute. \\
\\
I will provide you with images from two different viewpoints: \\
1. HEAD CAMERA: provides a head-level (human-like) viewpoint, showing the entire scene from the robot's eye height. This viewpoint is used to locate containers. \\
2. WRIST CAMERA: mounted on the arm's end effector, moves with the arm in real time, providing a close-up view for precise manipulation. \\
\\
Use the exact English object names from this list: \{allow\_entities\} \\
Your task is: \{instruction\} \\
\\
After each action, you will receive a new observation image reflecting the updated state. To complete the task, you need to output your action from the available action library. \\
\\
Available actions: \\
- pick \textless{}object\textgreater{}: move from the current position to a suitable position to grasp the target object, then close the gripper to pick it up. \\
- place to \textless{}container\textgreater{}: move from the current position to a suitable position to place the target object into the target container, then open the gripper to release it. \\
- lift: raise the end effector vertically. \\
- pull: withdraw the hand horizontally while keeping the gripper state unchanged. \\
- push: move the end effector inward while keeping the gripper state unchanged. \\
- observe: reset the end effector to the default position to obtain a clear, wide-angle view, or stabilize the camera orientation before executing navigation skills. \\
- open\_door \textless{}container\textgreater{}: with the door handle already grasped, swing the door open along an arc around the door axis, then release the handle. When executing, output the bounding box of the entire target container in the head camera image. \\
- open\_sink \textless{}sink\textgreater{}: with the faucet handle already grasped, lift the handle vertically to turn on the water flow. When executing, output the bounding box of the entire sink in the head camera image. \\
- close\_door: Push or pull the door along its path until fully closed. \\
- recall \textless{}step\_x\textgreater{}: if you are unsure whether a previous step (e.g. step\_0) succeeded, review the keyframe of that step. \\
- rotate\_knob: rotate the gripper without changing the current grasp position, usually used to turn a knob. \\
- end: if you believe the task is complete, output `end'. \\
\\
Follow these output rules: \\
- At each step, output one line of action and one line of reasoning (briefly explain why this action and target were chosen). If an action later turns out to have failed (e.g. the object was not picked up after pick), you may retry. \\
- pick, place, open\_door, and open\_sink additionally require a bounding box normalized to 0-1000, in the format: action: \textless{}action\textgreater{}, box\_2d: [y\_min, x\_min, y\_max, x\_max], where: \\
  - pick: the functional grasp region of the target object in the wrist camera image (e.g. handle, knob, or body center), avoiding non-graspable protruding parts; \\
  - place: the interior or receiving area of the target container in the head camera image, avoiding handles, spouts, and other non-receiving parts; \\
  - open\_door: the bounding box of the entire target container in the head camera image, with the target container specified in the action (e.g. open\_door fridge); \\
  - open\_sink: the bounding box of the entire sink in the head camera image, with the target sink specified in the action (e.g. open\_sink sink); \\
- recall: output in two lines, in the format: \\
  action: recall \\
  target: step\_x \\
\\
- All other actions (e.g. lift, pull) output only the action itself: action: \textless{}action\_name\textgreater{}. \\
- Do not include image data or base64 strings in the output. \\
\\
Examples: \\
- Case 1 (action with bounding box): \\
action: pick handle, box\_2d: [y\_min, x\_min, y\_max, x\_max] \\
reasoning: your reasoning \\
\\
- Case 2 (action without bounding box): \\
action: lift \\
reasoning: your reasoning \\
\\
- Case 3 (action with bounding box): \\
action: place to plate, box\_2d: [y\_min, x\_min, y\_max, x\_max] \\
reasoning: your reasoning \\
\\
- Case 4 (recall action): \\
action: recall \\
target: step\_0 \\
reasoning: your reasoning \\
\\
- Case 5 (open\_door action with bounding box): \\
action: open\_door fridge, box\_2d: [y\_min, x\_min, y\_max, x\_max] \\
reasoning: your reasoning \\
\\
- Case 6 (open\_sink action with bounding box): \\
action: open\_sink sink, box\_2d: [y\_min, x\_min, y\_max, x\_max] \\
reasoning: your reasoning
\end{tcolorbox}

\begin{tcolorbox}[
    enhanced,
    breakable,                      
    boxrule=1.5pt,                  
    colframe=brown!80!black,        
    colback=blue!5!white,           
    colbacktitle=brown!70!black,    
    title={Mobile Navigation System Prompt without BEV}, 
    fonttitle=\large\bfseries\sffamily, 
    arc=6pt,                        
    sharp corners=south,            
]
\textbf{System Prompt:} \\
\small 
You are a robotic arm composed of seven links with a white movable chassis in given room. You need to complete the tasks according to human instructions. \\
We provide an Available\_Actions set and the corresponding explanations for each action. Each step, you should select one action from Available\_Actions. \\
\\
I will provide you with images from two different ego-perspectives: \\
1. HEAD CAMERA: A high-angle perspective from behind and above the robot, looking forward. You can see the wide and high-level view of the environment ahead. Use this as your PRIMARY navigation, target searching, obstacle detection, and placing reference. \\
2. WRIST CAMERA: A close-up view from the end effector. Use this for precise grasping and assisting in verification. \\
\\
This is your task: \{instruction\} \\
Use the exact object names from this list as reference: \{allow\_entities\} \\
\\
After executing the action, you will receive a new observation image showing the updated state. \\
\\
NAVIGATION ACTIONS: \\
- move\_forward: Move forward a specific distance in the current viewing direction. Note: Executing this action will AUTOMATICALLY reset the head camera to look straight ahead (look\_forward) before moving. \\
- rotate\_right: 90° clockwise base rotation. Note: Executing this action will AUTOMATICALLY reset the head camera to look straight ahead (look\_forward). \\
- rotate\_left: 90° counter-clockwise base rotation. Note: Executing this action will AUTOMATICALLY reset the head camera to look straight ahead (look\_forward). \\
\\
MANIPULATION ACTIONS: \\
- pick \textless{}object\textgreater{}: Move end effector to grasp the target object and close the gripper. \\
- place to \textless{}container\textgreater{}: Move end effector to the target container and open the gripper to place. \\
- lift: Lift the end effector vertically. \\
- pull: withdraw the hand horizontally while keeping the gripper state unchanged. \\
- push: move the end effector inward while keeping the gripper state unchanged. \\
- observe: Reset the end-effector to the default home position for a clear view and stable robot state. \\
- open\_door \textless{}container\textgreater{}: with the door handle already grasped, swing the door open along an arc around the door axis, then release the handle. When executing, output the bounding box of the entire target container in the head camera image. \\
- open\_sink \textless{}sink\textgreater{}: with the faucet handle already grasped, lift the handle vertically to turn on the water flow. When executing, output the bounding box of the entire sink in the head camera image. \\
- rotate\_knob: rotate the gripper without changing the current grasp position, usually used to turn a knob. \\
- close\_door: Close the door. Automatically moves the end effector to the open door and performs the closing motion. \\
- recall \textless{}step\_x\textgreater{}: Review the key frames of a previous step (e.g. step\_0) if you are unsure whether it succeeded. \\
- end: Task completed. \\
\\
OUTPUT RULES AND FORMAT: \\
for every step, you must provide your decision and reasoning. the 2d bounding box [ymin, xmin, ymax, xmax] (normalized 0-1000) is required for pick, place, open\_door, and open\_sink actions. \\
for pick, target the functional grasping area of the object (e.g., handle, knob, or body center), avoiding non-graspable protrusions. for place, target the interior or intended receiving region of the container, excluding handles, spouts, or other non-receiving parts. for open\_door, output the bounding box of the entire target container in the head camera image. for open\_sink, output the bounding box of the entire sink in the head camera image. \\
\\
your output must strictly follow these formats: \\
\\
action: \textless{}action\_name\textgreater{} \\
reasoning: your reasoning \\
\\
- for pick action: \\
action: pick \textless{}object|knob|handle\textgreater{} \\
cam\_view: \textless{}wrist|head\textgreater{} \\
box\_2d: [ymin, xmin, ymax, xmax] \\
reasoning: your reasoning \\
\\
- for place action: \\
action: place to \textless{}container\textgreater{} \\
box\_2d: [ymin, xmin, ymax, xmax] \\
reasoning: your reasoning \\
\\
- for open\_door action: \\
action: open\_door \textless{}container\textgreater{} \\
box\_2d: [ymin, xmin, ymax, xmax] \\
reasoning: your reasoning \\
\\
- for open\_sink action: \\
action: open\_sink \textless{}sink\textgreater{} \\
box\_2d: [ymin, xmin, ymax, xmax] \\
reasoning: your reasoning \\
\\
examples: \\
\\
action: rotate\_left \\
reasoning: your reasoning \\
\\
action: move\_forward \\
reasoning: your reasoning \\
\\
action: pick apple \\
cam\_view: wrist \\
box\_2d: [300, 200, 500, 400] \\
reasoning: your reasoning \\
\\
action: place to plate \\
box\_2d: [200, 150, 400, 350] \\
reasoning: your reasoning \\
\\
action: open\_door fridge \\
box\_2d: [200, 150, 400, 350] \\
reasoning: your reasoning \\
\\
action: open\_sink sink \\
box\_2d: [250, 200, 450, 400] \\
reasoning: your reasoning \\
\\
action: rotate\_knob \\
reasoning: your reasoning \\
\\
action: end \\
reasoning: your reasoning \\
\\
action: recall \\
target: step\_0 \\
reasoning: your reasoning \\
\\
TASK PHASES AND STRATEGY: \\
This task requires alternating between navigation and manipulation actions. Execute in the cycle of ``navigate to the target area $\rightarrow$ perform manipulation'' until the task is completed. Do not navigate without performing operations. \\
Dynamically adjust your strategy based on real-time visual feedback from all views: \\
\\
1. Perform the ``rotate'' skill to observe, if the target is still not found, then execute move\_forward to change your physical position, clear visual blind spots, and explore new areas. \\
2. The target is still at a distance: in this phase, prioritize move\_forward to close the gap. If the target is off to the side, do not rush to rotate toward it --- with 90° rotation precision, if you and the target are at diagonal corners of the room, rotating in place will only cause the target to jump between the left and right sides of your view or disappear entirely, never achieving proper alignment. If there is an obstacle ahead, use rotate combined with move\_forward to detour; otherwise, continue moving forward. If moving forward repeatedly fails to bring you closer to the target, it indicates an obstacle is blocking the path.
\end{tcolorbox}

The operational context advances through a rolling interleaved history formatted to capture sequential execution steps ($n$). At the initial step ($n=0$), the model is provided with the task-specific system prompt alongside raw initial observations. For subsequent rollouts ($n > 0$), the pipeline dynamically constructs the prompt using three core operational fields:

\noindent \textbf{Context Windowing}: The text history maintains the full sequence of past actions and bounding boxes, whereas continuous visual frames are managed via a sliding maximum-length window governed by \texttt{history\_maxlen}. Elements exceeding this boundary are truncated with an \textit{(observation omitted)} placeholder to mitigate context explosion.

\noindent \textbf{Attention Guidance}: When execution constraints are triggered by physical anomalies or misalignments, the framework injects targeted error-correction hints into the history string. If the agent initiates premature grasping outside the operational workspace or fails viewport verification, the context automatically appends explicit attention feedback tokens based on the current state.

\noindent \textbf{Current Observation}: Each query slice terminates with a standardized sensory update block containing the latest head and wrist (and optional BEV) camera views, forcing the model to generate the next action contingent on both the historical trace and immediate sensory states.

The comprehensive semantic templates and historical rollout structures for both manipulation and mobile navigation paradigms are formalized as follows.

\begin{tcolorbox}[
    enhanced,
    breakable,                      
    boxrule=1.5pt,                  
    colframe=brown!80!black,        
    colback=blue!5!white,           
    colbacktitle=brown!70!black,    
    title={Semantic Template: Manipulation Tasks},      
    fonttitle=\large\bfseries\sffamily, 
    arc=6pt,                        
    sharp corners=south,            
]

\textbf{System Prompt:} \\
\textit{Manipulation System Prompt} \\

\vspace{-0.7em} 
\noindent\rule{\textwidth}{0.5pt}
\vspace{-0.7em} 

\textbf{Initial Step ($n=0$):} \\
These are your initial observation. Please analyze the scene and select a manipulation action. \\
1. HEAD CAMERA: \textless{}image\_1\textgreater{} \\
2. WRIST CAMERA: \textless{}image\_2\textgreater{} \\

\vspace{-0.7em} 
\noindent\rule{\textwidth}{0.5pt}
\vspace{-0.7em} 

\textbf{Subsequent Rollouts ($n > 0$):} \\
Your historical actions and recent observations are as follows: \\
\textbf{step 1}: action:\textless{}action\_1\textgreater{}, bbox:\textless{}bbox\_1\textgreater{}, observation at this time: (observation omitted); \textit{\% Exceeding history\_maxlen} \\
\textbf{step 2}: action:\textless{}action\_2\textgreater{}, bbox:\textless{}bbox\_2\textgreater{}, observation at this time: \textless{}image\_1\textgreater{} \textless{}image\_2\textgreater{}; \\
\dots \\
\textbf{step n}: action:\textless{}action\_n\textgreater{}, bbox:\textless{}bbox\_n\textgreater{},  \\
attention: Your pick action for the target was not executed because the target object is not in the current field of wrist view or is too far away to grasp, observation at this time: \textless{}image\_m\textgreater{} \textless{}image\_m+1\textgreater{}; \textit{\% Out of operational range} \\
\\
Your current observations are as follows: \textless{}image\_k\textgreater{} \textless{}image\_k+1\textgreater{} \\
Please select a manipulation action or output `action: end' if you have finished the task.
\end{tcolorbox}

\begin{tcolorbox}[
    enhanced,
    breakable,                      
    boxrule=1.5pt,                  
    colframe=brown!80!black,        
    colback=blue!5!white,           
    colbacktitle=brown!70!black,    
    title={Semantic Template: Mobile Navigation Tasks}, 
    fonttitle=\large\bfseries\sffamily, 
    arc=6pt,                        
    sharp corners=south,            
]
\textbf{System Prompt:} \\
\textit{Mobile Navigation System Prompt (with/without BEV)} \\

\vspace{-0.7em} 
\noindent\rule{\textwidth}{0.5pt}
\vspace{-0.7em} 

\textbf{Initial Step ($n=0$):} \\
These are your initial observation. This is a composite task requiring navigation followed by manipulation. \\
Phase 1 - Navigation: First navigate to the target location and adjust your orientation. \\
Phase 2 - Manipulation: Only after reaching the target position, execute manipulation skills. \\
\textit{\% Sensory inputs scale dynamically based on configuration (e.g., 2 views or 3 views including BEV)} \\

\vspace{-0.7em} 
\noindent\rule{\textwidth}{0.5pt}
\vspace{-0.7em} 

\textbf{Subsequent Rollouts ($n > 0$ 3-View Example):} \\
Your historical actions and recent observations are as follows: \\
\textbf{step 1}: action:\textless{}action\_1\textgreater{}, bbox:\textless{}bbox\_1\textgreater{}, observation at this time: (observation omitted); \\
\textbf{step 2}: action:\textless{}action\_2\textgreater{}, bbox:\textless{}bbox\_2\textgreater{}, observation at this time: \textless{}image\_1\textgreater{} \textless{}image\_2\textgreater{} \textless{}image\_3\textgreater{}; \\
\textbf{step 3}: action:\textless{}action\_3\textgreater{}, bbox:\textless{}bbox\_3\textgreater{}, attention: You are still in the navigation phase. The robot cannot reach the target object from the current location., observation at this time: \dots; \\
\textbf{step 4}: action:\textless{}action\_4\textgreater{}, bbox:\textless{}bbox\_4\textgreater{},  \\
attention: Your pick action for target object was not executed because the target object is not in the current field of wrist view or is too far away to grasp., observation at this time: \dots; \\
\textbf{step 5}: action:\textless{}action\_5\textgreater{}, bbox:\textless{}bbox\_5\textgreater{}, observation at this time: \dots; \\
\dots \\
\\
Your current observations are as follows: \textless{}image\_m\textgreater{} \textless{}image\_m+1\textgreater{} \textless{}image\_m+2\textgreater{} \\
Please continue navigating or output `action: end' if you have finished the task.
\end{tcolorbox}

\section{Qualitative Analysis and Failure Modes}
\label{app:qualitative}

\begin{figure}[!htbp]
\centering
\includegraphics[width=1\textwidth]{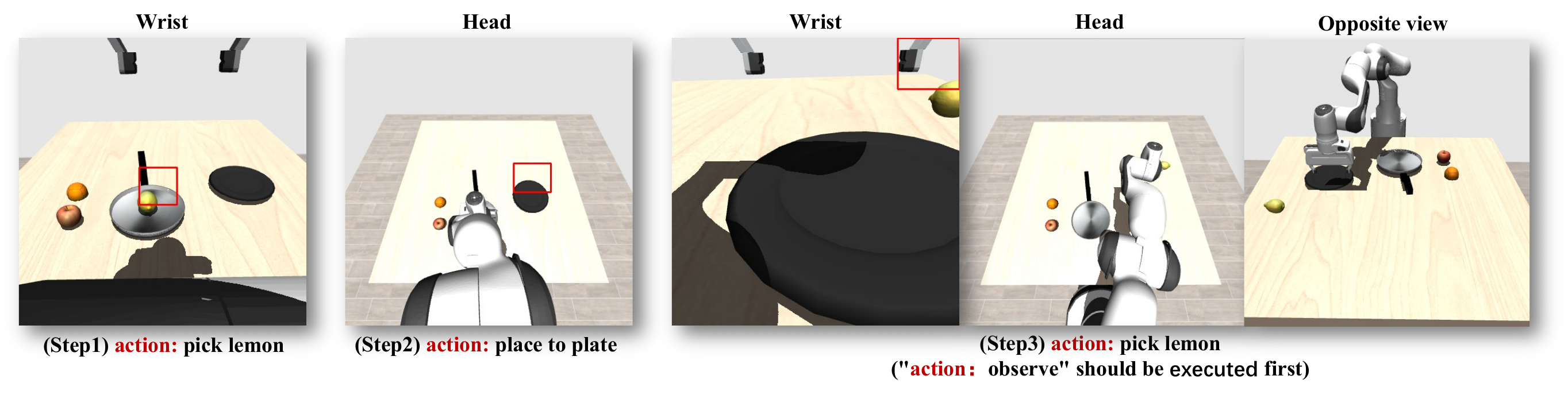}
\caption{An example of execution blindness and viewpoint adjustment failure by SEED-2.0 during the ``move the lemon from the pot to the bowl'' task.}
\label{fig:picklemon}
\end{figure}

\begin{figure}[t]
\centering
\includegraphics[width=1\textwidth]{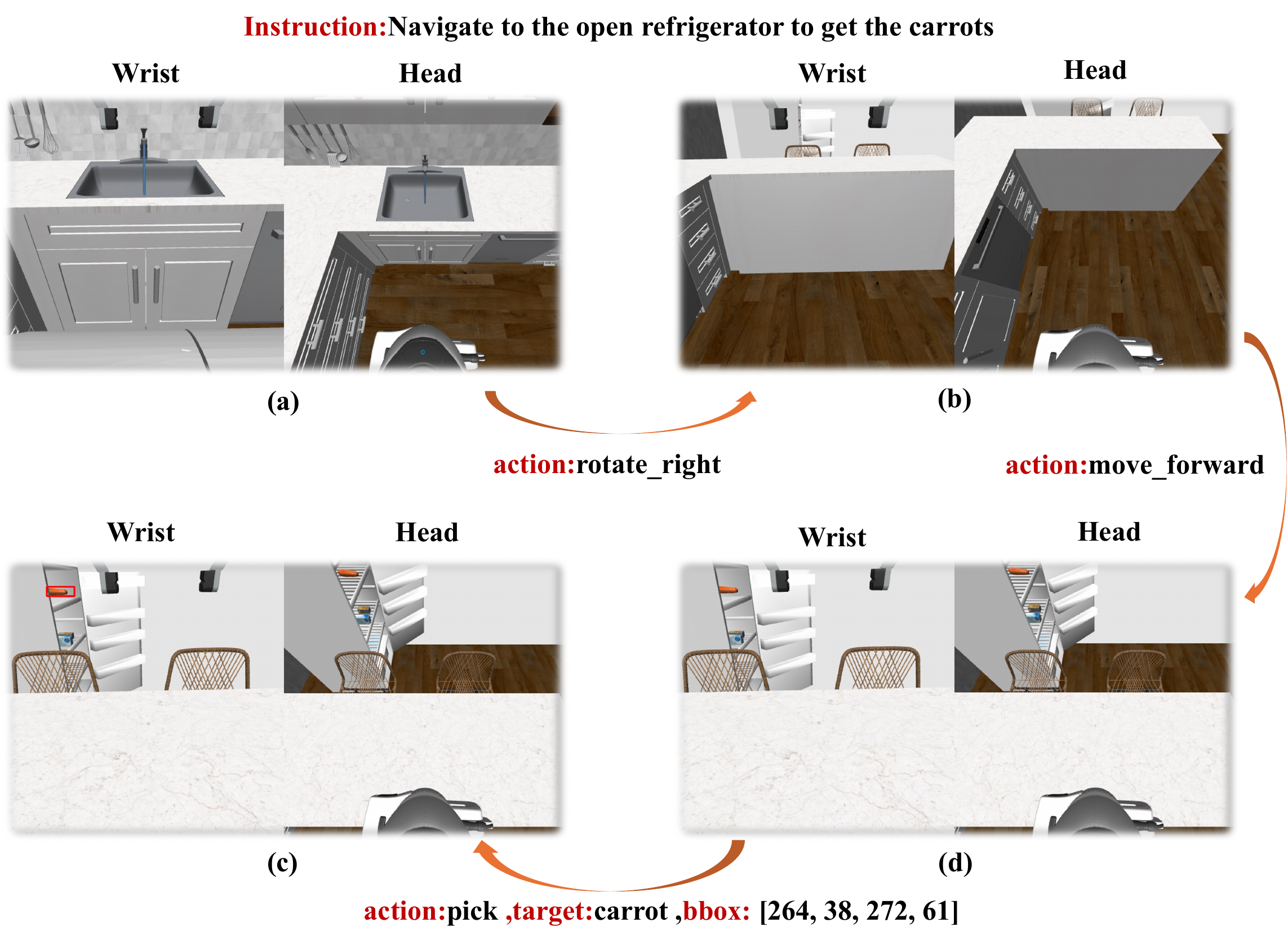}
\caption{An example of a temporal-spatial reasoning deficit. The model fails to utilize historical visual context to maintain obstacle awareness once local 3D boundary features degrade.}
\label{fig:pickcarrot}
\end{figure}

\textbf{Action Rigidity and Execution Blindness.} Although the models exhibit reasonable proficiency in high-level common-sense planning, they show a tendency to repeat identical action primitives rather than dynamically adjusting when localized execution failures occur. For instance, as shown in \Cref{fig:picklemon}, during the task ``move the lemon from the pot to the bowl'' executed by SEED-2.0, the model issued a place command, but the lemon failed to enter the bowl. Instead of calling an exploratory primitive like observe to regain a wide view and replan, the model directly repeated the pick command. Since the object was still within the field of view and recognized by the VLM, the model misjudged the current perspective as sufficient, which prompted it to directly repeat the pick command instead of invoking an observe action to reset its view. However, as a result of consecutive pick attempts, the wrist camera moved too close to the target. This narrow field of view led to highly inaccurate predictions of the bounding box. Since the system completely relies on bounding boxes for closed-loop grasping, blindly repeating pick under limited visibility and coordinate drift caused the robot to fall into a failure loop, revealing the model's insufficiency in actively adjusting its viewpoint when local visibility degrades.

\textbf{Temporal-Spatial Reasoning Deficit.} In composite navigation-manipulation tasks, the models' capability to utilize historical image sequences to maintain current spatial awareness is fragile. For example, as shown in \Cref{fig:pickcarrot}, during the task ``navigate to the open refrigerator to get the carrots'' executed by Gemini-3-Flash, the white table obstacle ahead is clearly visible in the distant view during the early phase (e.g., in step a). However, as the mobile base executes the rotate\_right and move\_forward commands to approach the workstation (moving from step b to c), the white table loses its 3D boundary features in the head camera's view, showing only a plain white surface. At this point, the model fails to utilize the distant visual memory from the sequence history to infer that the table still acts as an impassable obstacle ahead. Instead of adopting a strategy to bypass the obstacle, it continues to move forward and prematurely attempts to issue the local manipulation command pick with a target bounding box (step c and d). This failure in temporal-spatial reasoning eventually causes the robot to continuously collide forward into the obstacle and fail the grasp due to its inability to reach the target.

\begin{figure}[!htbp]
\centering
\includegraphics[width=1\textwidth]{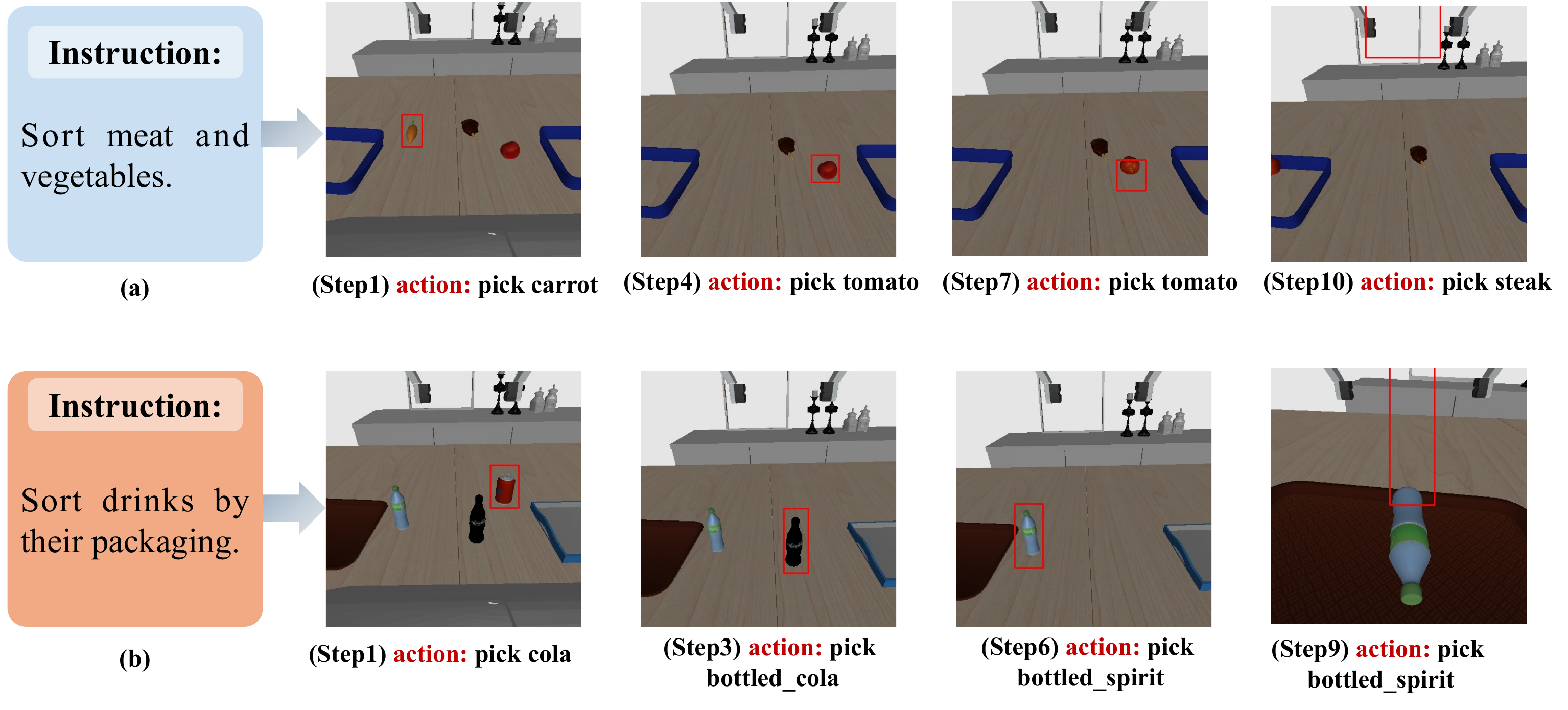}
\caption{Examples of spatial grounding degradation over long execution steps, with both rollouts executed by SEED-2.0; intermediate place and observe skills are omitted. The model provides precise BBoxes in the early steps but suffers from severe coordinate hallucination and upward drift beyond Step 8.}
\label{fig:grounding_degradation}
\end{figure}

\subsection{Spatial Grounding Degradation in Long Sequences}
To analyze the stability of 2D spatial grounding as interaction depth increases, we first examine a representative case and then verify its underlying cause through a controlled experiment. Note that the sequences in \Cref{fig:grounding_degradation} only capture the wrist camera's perspective and the corresponding object BBoxes during the execution of the pick skill, omitting the intermediate place and observe skills for brevity. As illustrated in \Cref{fig:grounding_degradation}, which showcases two long-horizon tasks executed by SEED-2.0, the spatial grounding performance maintains high precision during the early steps of execution. In task (a), ``Sort meat and vegetables'', the BBoxes tightly bound the carrot and tomato from Step 1 to Step 7. Similarly, in task (b), ``Sort drinks by their packaging'', the model accurately locates the cola can, bottled cola, and the light-blue bottled drink from Step 1 to Step 6. However, as the task progresses beyond Step 8 and historical frames accumulate in the context window, a severe deterioration in spatial grounding becomes evident. By Step 10 in task (a) and Step 9 in task (b), the predicted BBoxes for the steak and the light-blue bottled drink exhibit extreme coordinate hallucinations, drifting completely away from the target objects and floating into the upper empty space of the frame, directly triggering critical grasping failures in the later phases of long-horizon tasks. This observation is consistent with the hypothesis that accumulated context attenuates the pixel-wise spatial fidelity of VLMs while largely preserving high-level semantic action selection.

To disentangle the relationship between context length and localization accuracy, this subsection provides the full details of the controlled experiment summarized in the main paper. We fix the visual input, vary only the number of history frames in the context (2--16), and evaluate the BBox IoU of each model under two conditions: a normal scene and a long-distance, occluded scene.

\begin{table*}[t]
\centering
\caption{BBox IoU vs. visual context frames (2--16), 10 runs per scene.}
\vspace{2pt}
\label{tab:iou_detail}
\renewcommand{\arraystretch}{1.2}
\setlength{\tabcolsep}{4pt}
\resizebox{\textwidth}{!}{
\begin{tabular}{l|lcccccccc}
\toprule
\textbf{Model} & \textbf{Cond.} & \textbf{2} & \textbf{4} & \textbf{6} & \textbf{8} & \textbf{10} & \textbf{12} & \textbf{14} & \textbf{16} \\
\midrule
\multirow{2}{*}{{Gemini-3-Pro}}& Normal & 0.73$\pm$0.01 & 0.73$\pm$0.01 & 0.72$\pm$0.01 & 0.65$\pm$0.23 & 0.72$\pm$0.03 & 0.66$\pm$0.23 & 0.73$\pm$0.01 & 0.65$\pm$0.23 \\
               & Ld+Occ & 0.80$\pm$0.02 & 0.56$\pm$0.39 & 0.55$\pm$0.38 & 0.55$\pm$0.38 & 0.64$\pm$0.34 & 0.16$\pm$0.34 & 0.39$\pm$0.42 & 0.32$\pm$0.41 \\
\midrule
\multirow{2}{*}{{Gemini-3-Flash}}& Normal & 0.71$\pm$0.01 & 0.72$\pm$0.01 & 0.72$\pm$0.01 & 0.72$\pm$0.01 & 0.71$\pm$0.01 & 0.72$\pm$0.01 & 0.73$\pm$0.01 & 0.73$\pm$0.01 \\
               & Ld+Occ & 0.77$\pm$0.06 & 0.81$\pm$0.02 & 0.81$\pm$0.01 & 0.81$\pm$0.01 & 0.81$\pm$0.01 & 0.63$\pm$0.34 & 0.66$\pm$0.26 & 0.76$\pm$0.10 \\
\midrule
\multirow{2}{*}{{Qwen-3.5-Plus}} & Normal & 0.33$\pm$0.01 & 0.43$\pm$0.05 & 0.45$\pm$0.15 & 0.40$\pm$0.12 & 0.29$\pm$0.04 & 0.45$\pm$0.16 & 0.40$\pm$0.13 & 0.40$\pm$0.13 \\
               & Ld+Occ & 0.26$\pm$0.16 & 0.38$\pm$0.24 & 0.31$\pm$0.29 & 0.39$\pm$0.37 & 0.44$\pm$0.31 & 0.41$\pm$0.28 & 0.32$\pm$0.31 & 0.37$\pm$0.36 \\
\midrule
\multirow{2}{*}{{Doubao-SEED-2.0-Pro}} & Normal & 0.60$\pm$0.08 & 0.58$\pm$0.06 & 0.55$\pm$0.06 & 0.53$\pm$0.06 & 0.20$\pm$0.27 & 0.46$\pm$0.17 & 0.12$\pm$0.21 & 0.00$\pm$0.00 \\
               & Ld+Occ & 0.33$\pm$0.14 & 0.20$\pm$0.17 & 0.13$\pm$0.15 & 0.16$\pm$0.17 & 0.01$\pm$0.02 & 0.00$\pm$0.00 & 0.00$\pm$0.00 & 0.00$\pm$0.00 \\
\midrule
\multirow{2}{*}{{Gemini-2.5-Pro}} & Normal & 0.52$\pm$0.30 & 0.62$\pm$0.21 & 0.58$\pm$0.25 & 0.54$\pm$0.31 & 0.52$\pm$0.31 & 0.58$\pm$0.25 & 0.62$\pm$0.21 & 0.54$\pm$0.30 \\
               & Ld+Occ & 0.62$\pm$0.28 & 0.64$\pm$0.30 & 0.56$\pm$0.34 & 0.48$\pm$0.42 & 0.56$\pm$0.38 & 0.52$\pm$0.27 & 0.42$\pm$0.25 & 0.44$\pm$0.39 \\
\bottomrule
\end{tabular}
}
\end{table*}

The results listed in Table~\ref{tab:iou_detail} reveal that localization degradation is not a simple linear effect of context length but instead varies in a clearly model-dependent manner: models such as Gemini-3-Flash remain largely stable under the normal condition, whereas others fluctuate, though only within a narrow, trendless range; only SEED-2.0, the model shown in \Cref{fig:grounding_degradation}, exhibits a clear persistent degradation as the context grows, which is further exacerbated and emerges earlier under the long-distance, occluded condition.

This indicates that context length itself is not the direct cause of localization degradation. The degradation is more likely jointly driven by the scene complexity (e.g., long distance and occlusion) and the disparity in visual robustness across models. For models sensitive to complex visual conditions, BBox accuracy degrades as the input image sequence grows longer, and grasp failures become correspondingly more frequent, whereas models with stronger visual robustness maintain stable low-level spatial resolution over long contexts.

\subsection{Qualitative Execution Examples}
\label{app:astra_examples}

\begin{figure}[!htbp]
\centering
\includegraphics[width=\textwidth]{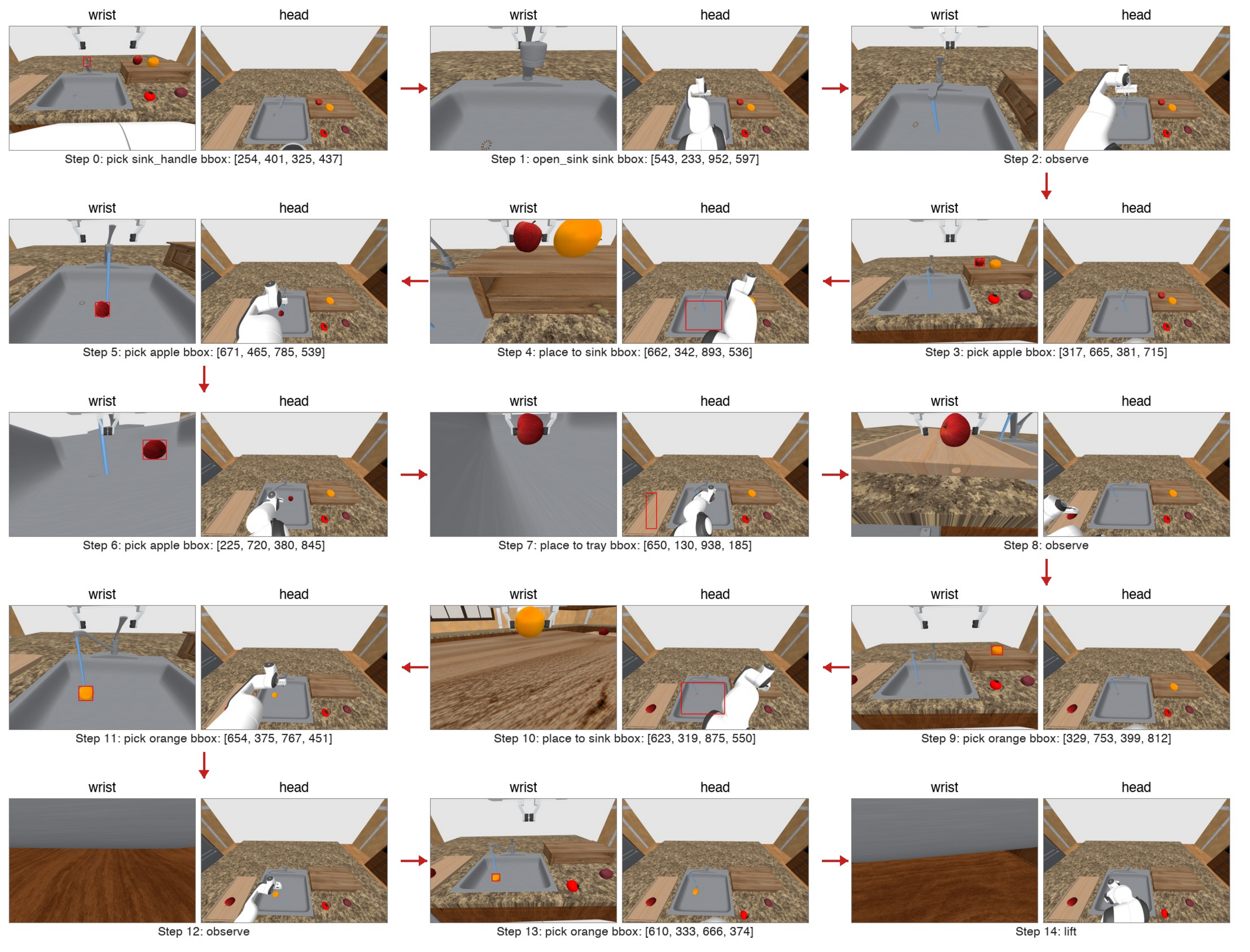}
\caption{``The fruit on the table shelf has not been washed. Wash every piece of fruit and return all of them to the tray.''}
\label{fig:astra_example}
\end{figure}

\begin{figure}[!htbp]
\centering
\includegraphics[width=\textwidth]{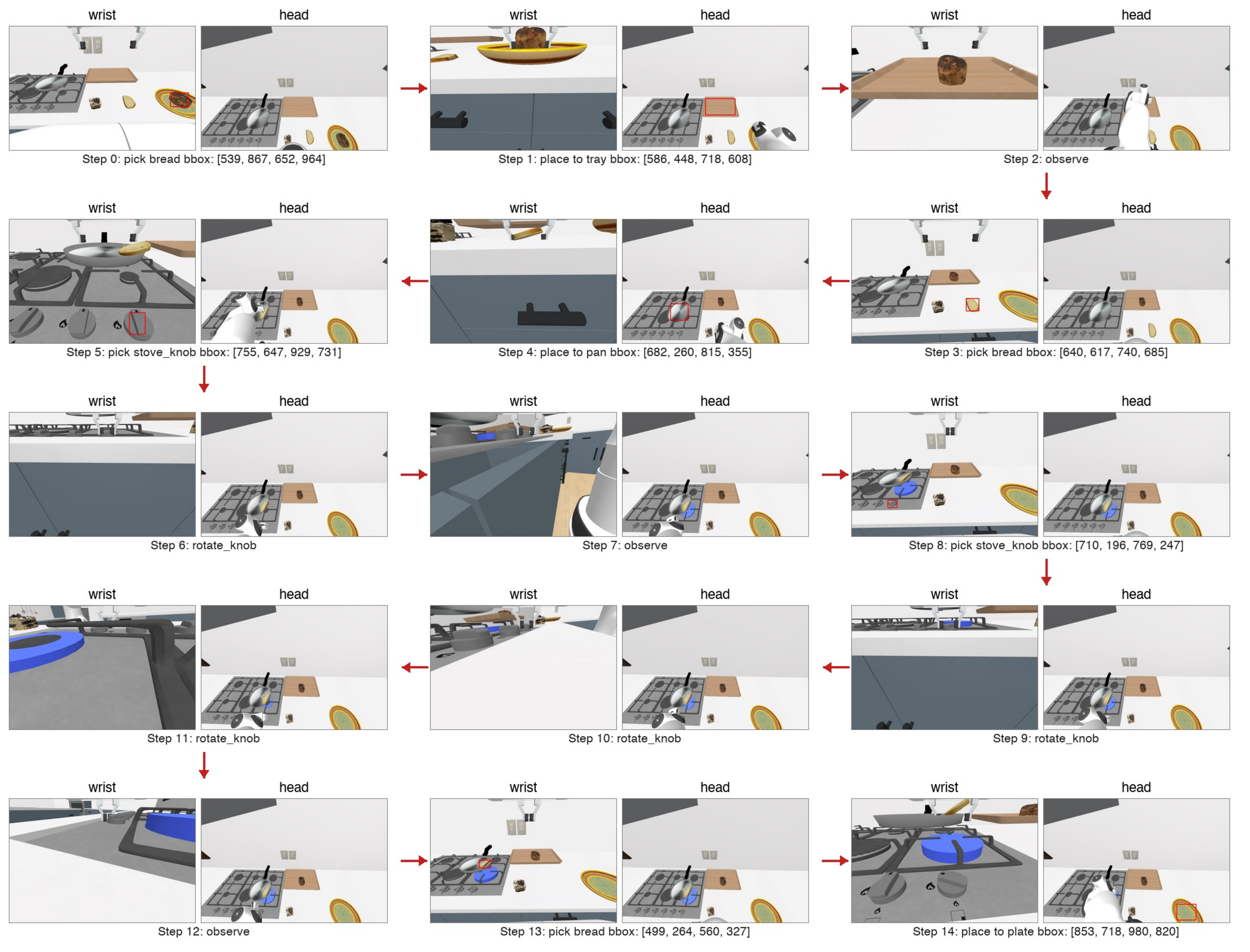}
\caption{``The bread on the plate is burnt, I would like it placed in the tray, and a fresh one pan-fried and placed on the same plate.''}
\label{fig:astra_fix_burnt_bread}
\end{figure}

\begin{figure}[!htbp]
\centering
\includegraphics[width=\textwidth]{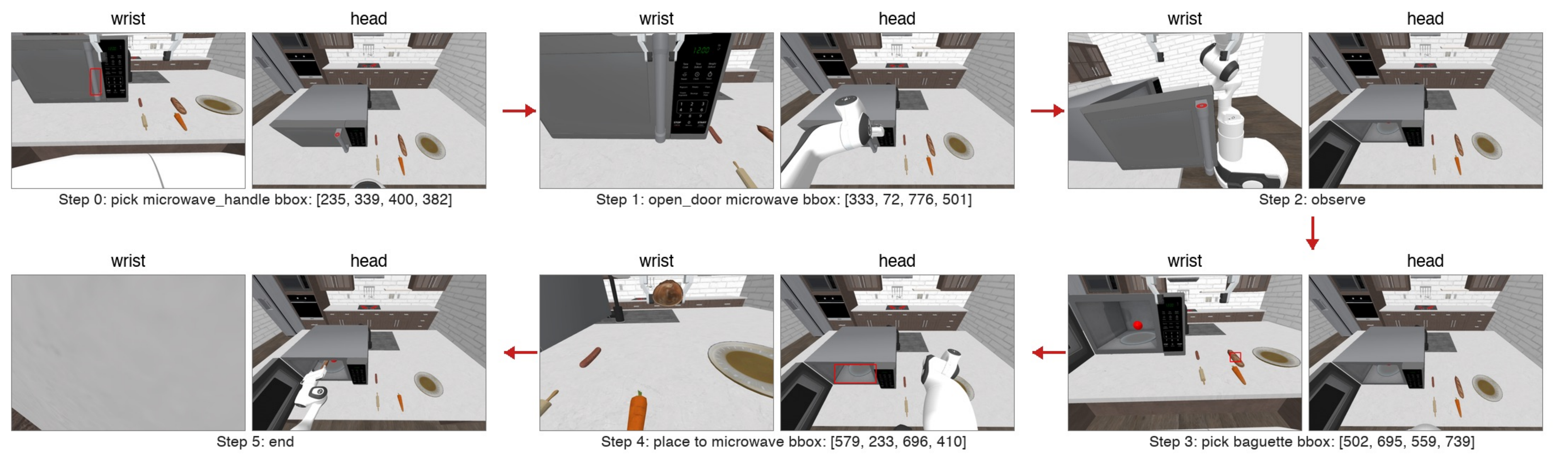}
\caption{``I want to eat the baguette, please warm it in the microwave.''}
\label{fig:astra_warm_baguette}
\end{figure}

\begin{figure}[!htbp]
\centering
\includegraphics[width=\textwidth,height=0.9\textheight,keepaspectratio]{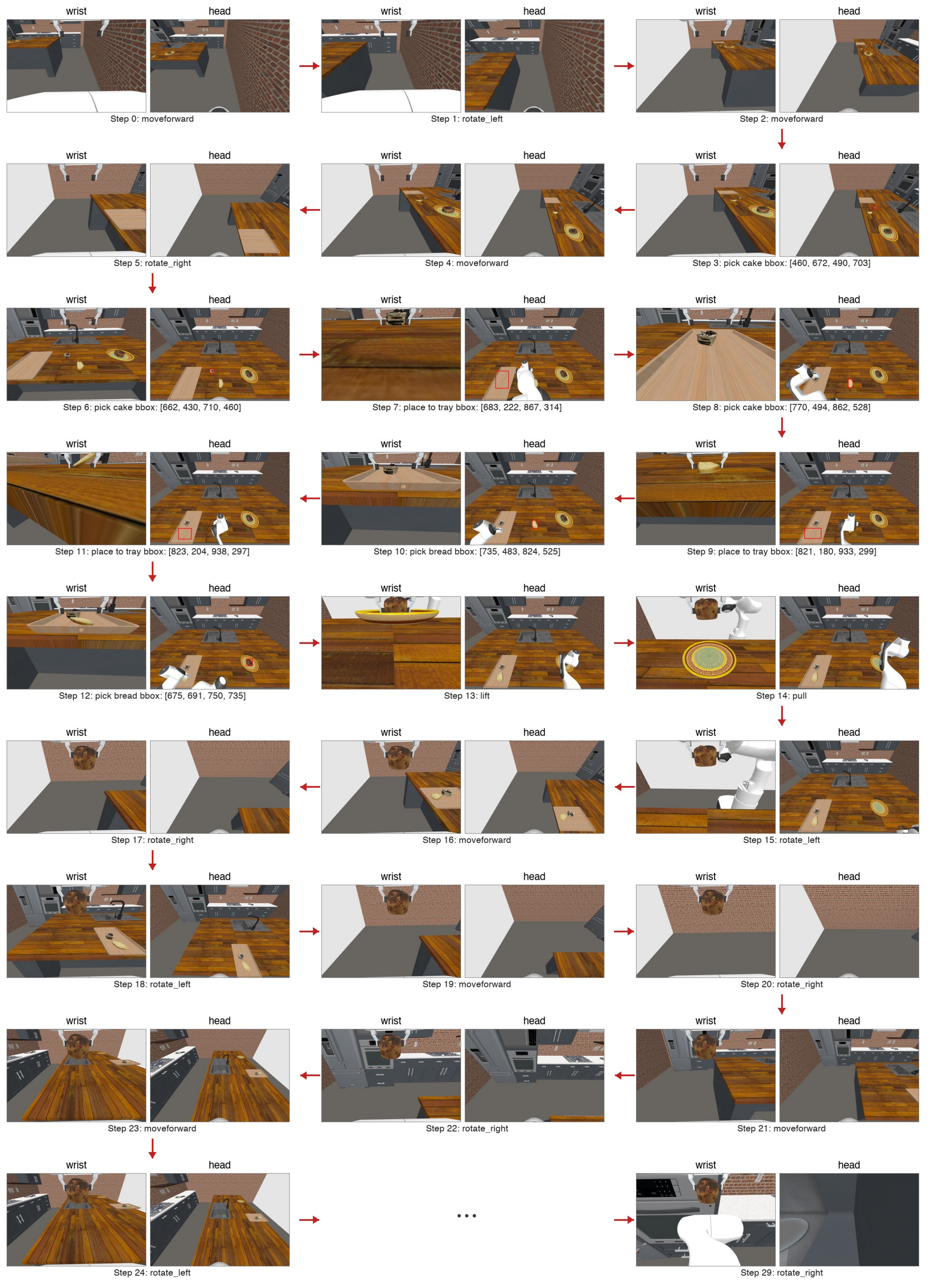}
\caption{``Navigate to help me put the rest of the food in the tray, then put the bread from the plate into the frying pan for preparation.''}
\label{fig:astra_nav_burnt_bread}
\end{figure}

\begin{figure}[!htbp]
\centering
\includegraphics[width=\textwidth,height=0.9\textheight,keepaspectratio]{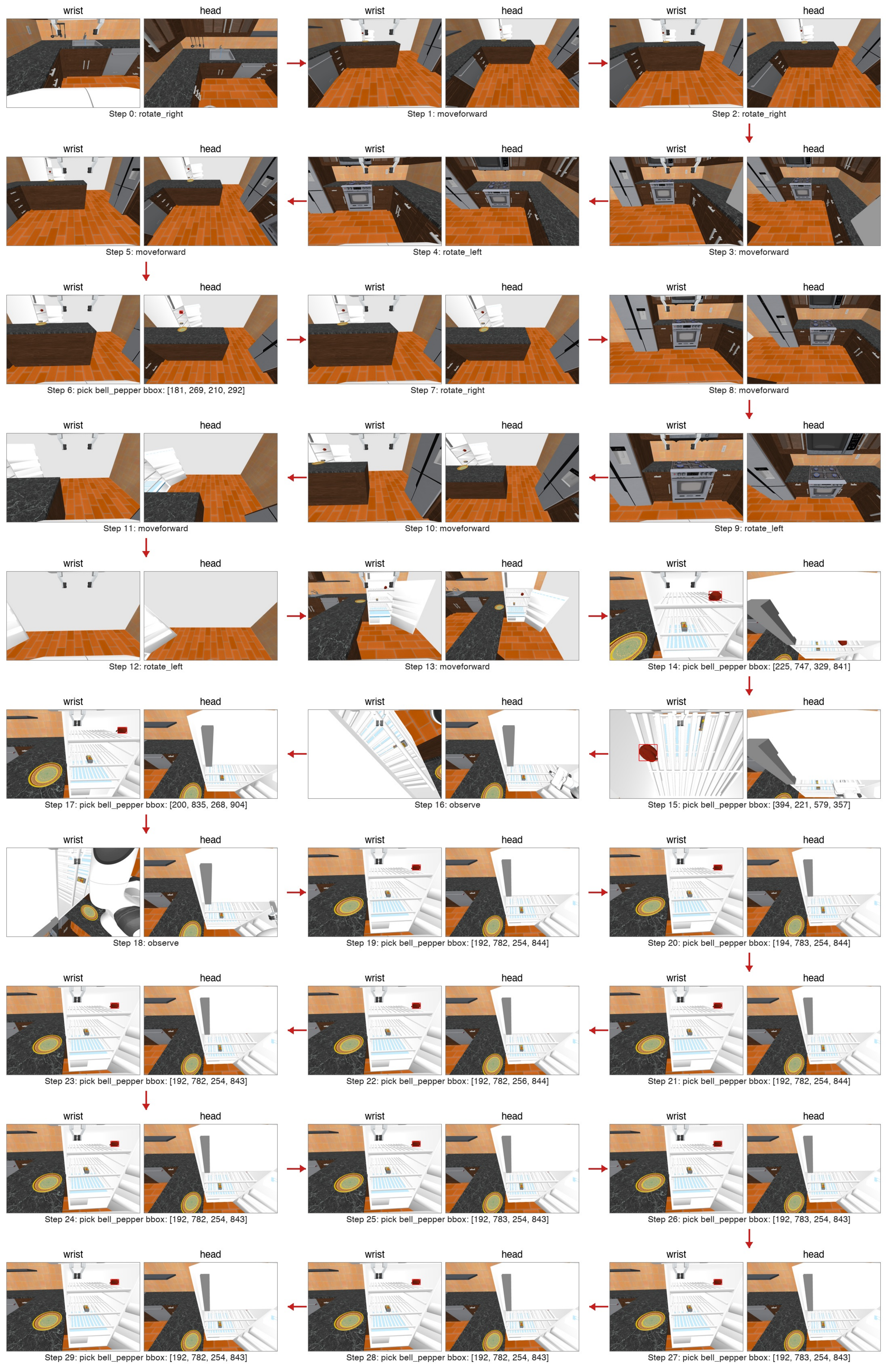}
\caption{``Navigate to the open main refrigerator and put the bell pepper into the plate next to it.''}
\label{fig:astra_nav_refrigerator}
\end{figure}

To complement the failure-mode analysis above, we present representative qualitative rollouts of GPT-6-Astra, the strongest-performing model in our evaluation, on both long-horizon and composite navigation tasks. These examples encompass both successful executions and attempts involving intermediate failures and subsequent retries, offering a more complete picture of how the model sequences navigation, grounding, and manipulation primitives in practice. For clarity, each figure is captioned directly with its task instruction, and the frames follow the rollout order indicated by the arrows, with the executed action and predicted BBox annotated beneath each step. Three representative long-horizon manipulation rollouts are shown in \Cref{fig:astra_example,fig:astra_fix_burnt_bread,fig:astra_warm_baguette}, covering fruit washing, pan-frying a fresh bread, and microwave reheating. In addition, two rollouts on composite navigation-manipulation tasks without BEV input are shown in \Cref{fig:astra_nav_burnt_bread,fig:astra_nav_refrigerator}.

\section{Full Task Configurations}
\begin{figure}[htbp]
    \centering
    \includegraphics[height=0.9\textheight]{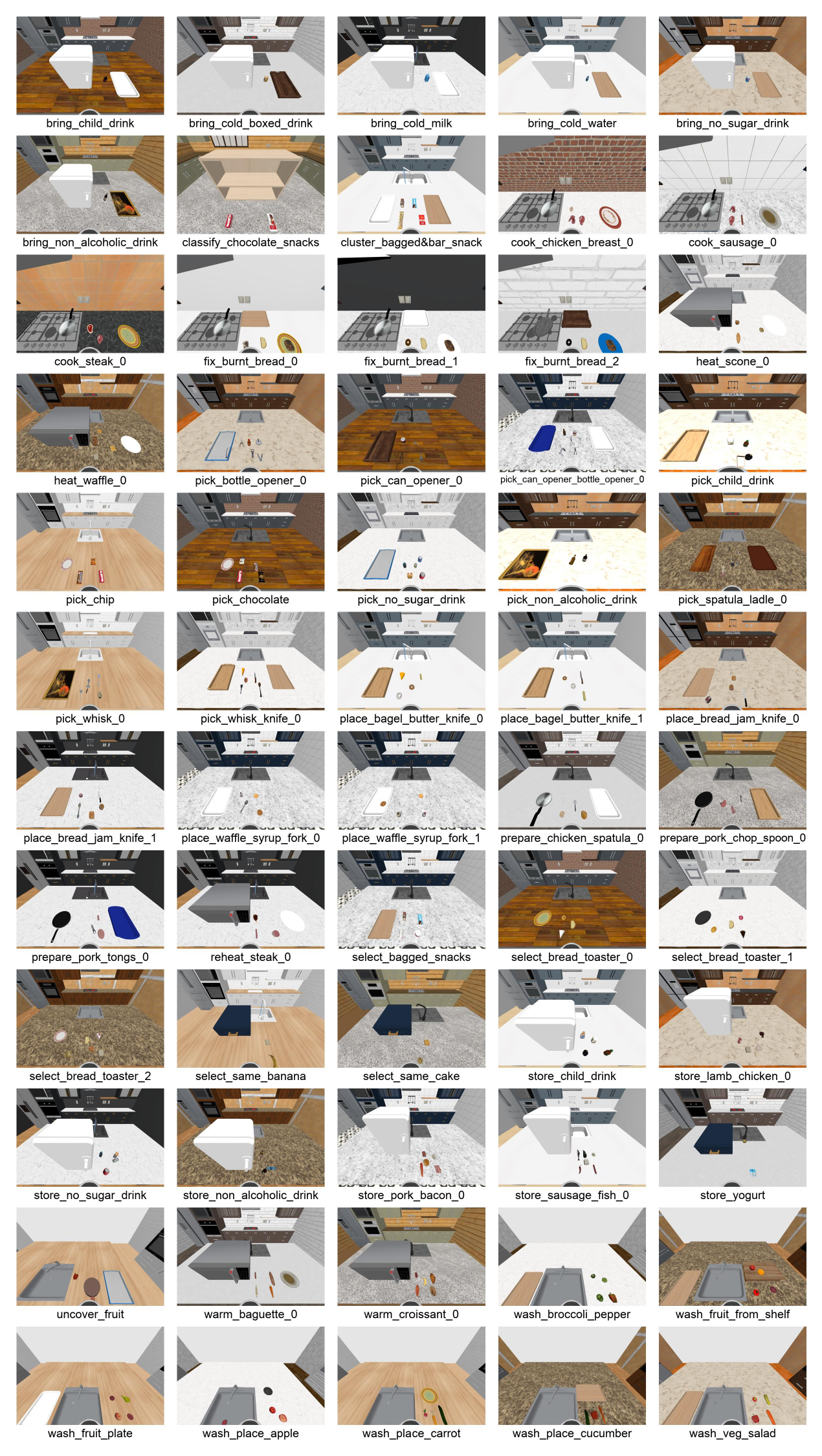}
    \caption{Head-camera views of the manipulation task configurations.}
    \label{fig:manip_visual}
\end{figure}

The composite navigation-manipulation tasks are constructed from the manipulation tasks: a single manipulation task can be decomposed into navigation and manipulation phases executed across different functional layouts, or multiple manipulation tasks located in different functional areas can be combined into one task. The instructions of all composite navigation-manipulation tasks are therefore derived from the manipulation task configurations listed in \Cref{tab:manip_tasks}. The head-camera views of all manipulation task configurations are shown in \Cref{fig:manip_visual}.

\clearpage
\begin{longtable}{@{}>{\raggedright\arraybackslash}p{0.26\textwidth}p{0.14\textwidth}p{0.545\textwidth}@{}}
\caption{Complete configuration of the 60 tabletop manipulation tasks.}
\label{tab:manip_tasks}\\
\toprule
\textbf{Task Name} & \textbf{Horizon} & \textbf{Instruction} \\
\midrule
\endfirsthead
\toprule
\textbf{Task Name} & \textbf{Horizon} & \textbf{Instruction} \\
\midrule
\endhead
\endfoot
\bottomrule
\endlastfoot
pick\_\allowbreak chocolate & Short-horizon & I want to eat the chocolate, can you put it on the plate for me? \\
\midrule
pick\_\allowbreak chip & Short-horizon & I want to eat the chips, can you put it on the plate for me? \\
\midrule
select\_\allowbreak bread\_\allowbreak toaster\_\allowbreak 0 & Short-horizon & I would like the bread on the plate. \\
\midrule
select\_\allowbreak bread\_\allowbreak toaster\_\allowbreak 1 & Short-horizon & I would like the bread on the plate. \\
\midrule
select\_\allowbreak bread\_\allowbreak toaster\_\allowbreak 2 & Short-horizon & I would like the bread for my sandwich on the plate. \\
\midrule
pick\_\allowbreak can\_\allowbreak opener\_\allowbreak 0 & Short-horizon & I would like to eat the food on the table, please choose a suitable tool and put it on the tray. \\
\midrule
pick\_\allowbreak whisk\_\allowbreak 0 & Short-horizon & I would like to cook the food on the table, please choose a suitable tool and put it on the tray. \\
\midrule
pick\_\allowbreak bottle\_\allowbreak opener\_\allowbreak 0 & Short-horizon & I would like to drink the beverage on the table, please choose a suitable tool and put it on the tray. \\
\midrule
wash\_\allowbreak place\_\allowbreak carrot & Long-horizon & I want to eat the carrot, please wash it and put it on the plate. \\
\midrule
wash\_\allowbreak place\_\allowbreak apple & Long-horizon & I want to eat the apple, please wash it and put it on the bowl. \\
\midrule
wash\_\allowbreak place\_\allowbreak cucumber & Long-horizon & I want to eat the cucumber, please wash it and put it in the tray. \\
\midrule
cook\_\allowbreak steak\_\allowbreak 0 & Long-horizon & I want to eat the steak, please cook it in the pan and put it on the plate. \\
\midrule
cook\_\allowbreak chicken\_\allowbreak breast\_\allowbreak 0 & Long-horizon & I want to eat the chicken breast, please cook it in the pan and put it on the plate. \\
\midrule
cook\_\allowbreak sausage\_\allowbreak 0 & Long-horizon & I want to eat the sausage, please cook it in the pan and put it on the plate. \\
\midrule
reheat\_\allowbreak steak\_\allowbreak 0 & Long-horizon & I want to eat the steak, please warm it in the microwave. \\
\midrule
heat\_\allowbreak scone\_\allowbreak 0 & Long-horizon & I want to eat the scone, please warm it in the microwave. \\
\midrule
heat\_\allowbreak waffle\_\allowbreak 0 & Long-horizon & I want to eat the waffle, please warm it in the microwave. \\
\midrule
warm\_\allowbreak baguette\_\allowbreak 0 & Short-horizon & I want to eat the baguette, please warm it in the microwave. \\
\midrule
warm\_\allowbreak croissant\_\allowbreak 0 & Short-horizon & I want to eat the croissant, please warm it in the microwave. \\
\midrule
store\_\allowbreak lamb\_\allowbreak chicken\_\allowbreak 0 & Short-horizon & I would like the lamb chop and the chicken breast stored in the refrigerator. \\
\midrule
store\_\allowbreak pork\_\allowbreak bacon\_\allowbreak 0 & Short-horizon & I would like the pork loin and the bacon stored in the refrigerator. \\
\midrule
store\_\allowbreak sausage\_\allowbreak fish\_\allowbreak 0 & Short-horizon & I would like the sausage and the fish stored in the refrigerator. \\
\midrule
bring\_\allowbreak cold\_\allowbreak water & Short-horizon & I would like a cold bottle of water. \\
\midrule
bring\_\allowbreak cold\_\allowbreak milk & Short-horizon & I would like some cold milk for breakfast. \\
\midrule
bring\_\allowbreak cold\_\allowbreak boxed\_\allowbreak drink & Short-horizon & I would like a cold boxed fruit drink. \\
\midrule
bring\_\allowbreak non\_\allowbreak alcoholic\_\allowbreak drink & Short-horizon & I am driving tonight, so I would like a non-alcoholic drink. \\
\midrule
bring\_\allowbreak child\_\allowbreak drink & Short-horizon & I would like a suitable drink for a child. \\
\midrule
bring\_\allowbreak no\_\allowbreak sugar\_\allowbreak drink & Short-horizon & I am trying to avoid sugary drinks, so I would like a suitable option. \\
\midrule
pick\_\allowbreak non\_\allowbreak alcoholic\_\allowbreak drink & Short-horizon & I am driving tonight, please recommend a drink for me on the tray. \\
\midrule
pick\_\allowbreak child\_\allowbreak drink & Short-horizon & Choose a suitable drink for a child. \\
\midrule
pick\_\allowbreak no\_\allowbreak sugar\_\allowbreak drink & Short-horizon & I'm trying to avoid sugary drinks. Bring me a suitable option. \\
\midrule
store\_\allowbreak non\_\allowbreak alcoholic\_\allowbreak drink & Short-horizon & I would like the non-alcoholic drink chilled. \\
\midrule
store\_\allowbreak child\_\allowbreak drink & Short-horizon & I would like a suitable drink for a child chilled. \\
\midrule
store\_\allowbreak no\_\allowbreak sugar\_\allowbreak drink & Short-horizon & I would like a no-sugar drink option chilled. \\
\midrule
fix\_\allowbreak burnt\_\allowbreak bread\_\allowbreak 0 & Long-horizon & The bread on the plate is burnt, I would like it placed in the tray, and a fresh one pan-fried and placed on the same plate. \\
\midrule
fix\_\allowbreak burnt\_\allowbreak bread\_\allowbreak 1 & Long-horizon & The bread on the plate is burnt, I would like it placed in the tray, and a fresh one pan-fried and placed on the same plate. \\
\midrule
fix\_\allowbreak burnt\_\allowbreak bread\_\allowbreak 2 & Long-horizon & The bread on the plate is burnt, I would like it placed in the tray, and a fresh one pan-fried and placed on the same plate. \\
\midrule
select\_\allowbreak same\_\allowbreak cake & Long-horizon & An object is already stored in the cabinet. Pick the object of the same kind on the table and put it into the cabinet. \\
\midrule
select\_\allowbreak same\_\allowbreak banana & Long-horizon & An object is already stored in the cabinet. Pick the object of the same kind on the table and put it into the cabinet. \\
\midrule
store\_\allowbreak yogurt & Long-horizon & Put the yogurt away into the cabinet. \\
\midrule
uncover\_\allowbreak fruit & Long-horizon & Help me wash the apple. \\
\midrule
select\_\allowbreak bagged\_\allowbreak snacks & Short-horizon & I'd like to have all the bagged snacks, can you put them on the tray for me? \\
\midrule
cluster\_\allowbreak bagged\&bar\_\allowbreak snack & Short-horizon & I want the bars and bagged foods sorted into two separate trays, can you help me with that? \\
\midrule
wash\_\allowbreak veg\_\allowbreak salad & Long-horizon & I want to make a vegetable salad, please wash the carrot, tomato, and cucumber and put them in the tray. \\
\midrule
wash\_\allowbreak fruit\_\allowbreak plate & Long-horizon & I want to eat the apple, orange, and pear, please wash them and put them on the plate. \\
\midrule
wash\_\allowbreak broccoli\_\allowbreak pepper & Long-horizon & I want to eat the broccoli and bell pepper, please wash them and put them in the tray. \\
\midrule
wash\_\allowbreak fruit\_\allowbreak from\_\allowbreak shelf & Long-horizon & The fruit on the table shelf has not been washed. Wash every piece of fruit and return all of them to the tray. \\
\midrule
prepare\_\allowbreak pork\_\allowbreak tongs\_\allowbreak 0 & Short-horizon & I would like the pork loin and the tongs prepared for pan-frying in the right spots. \\
\midrule
prepare\_\allowbreak chicken\_\allowbreak spatula\_\allowbreak 0 & Short-horizon & I would like the spatula and the chicken breast prepared for pan-frying in the right spots. \\
\midrule
prepare\_\allowbreak pork\_\allowbreak chop\_\allowbreak spoon\_\allowbreak 0 & Short-horizon & I would like the spoon and the pork chop prepared for pan-frying in the right spots. \\
\midrule
place\_\allowbreak bread\_\allowbreak jam\_\allowbreak knife\_\allowbreak 0 & Short-horizon & I would like the bread, the jam, and the knife on the preparation tray. \\
\midrule
place\_\allowbreak waffle\_\allowbreak syrup\_\allowbreak fork\_\allowbreak 0 & Short-horizon & I would like the waffle, the syrup, and the fork on the preparation tray. \\
\midrule
place\_\allowbreak bagel\_\allowbreak butter\_\allowbreak knife\_\allowbreak 0 & Short-horizon & I would like the bagel, the butter, and the knife on the preparation tray. \\
\midrule
place\_\allowbreak bread\_\allowbreak jam\_\allowbreak knife\_\allowbreak 1 & Short-horizon & I want to eat bread with jam, please put what I need on the preparation tray. \\
\midrule
place\_\allowbreak waffle\_\allowbreak syrup\_\allowbreak fork\_\allowbreak 1 & Short-horizon & My friend wants to eat waffle with syrup, please put the three things needed on the preparation tray. \\
\midrule
place\_\allowbreak bagel\_\allowbreak butter\_\allowbreak knife\_\allowbreak 1 & Short-horizon & I want to eat a buttered bagel for breakfast, please put what I need on the preparation tray. \\
\midrule
pick\_\allowbreak can\_\allowbreak opener\_\allowbreak bottle\_\allowbreak opener\_\allowbreak 0 & Short-horizon & I would like to eat the canned food and drink the beer, please put the suitable tools on the trays beside them respectively. \\
\midrule
pick\_\allowbreak whisk\_\allowbreak knife\_\allowbreak 0 & Short-horizon & I would like to cook the egg and cut the cheese, please put the suitable tools on the trays beside them respectively. \\
\midrule
pick\_\allowbreak spatula\_\allowbreak ladle\_\allowbreak 0 & Short-horizon & I would like to fry the food and serve the soup, please put the suitable tools on the trays beside them respectively. \\
\midrule
classify\_\allowbreak chocolate\_\allowbreak snacks & Long-horizon & Put chocolate-related snacks on the first layer and other snacks on the second layer. \\
\end{longtable}

\end{document}